\documentclass[a4paper]{article}

\usepackage{INTERSPEECH2021}

\usepackage{cite}
\usepackage{amsmath,amssymb,amsfonts}
\usepackage{algorithmic}
\usepackage{graphicx}
\usepackage{textcomp}
\usepackage{multirow}
\usepackage{hrefhide}
\usepackage{mathtools}

\usepackage{color, colortbl}
\definecolor{LightCyan}{rgb}{0.88,1,1}

\usepackage[first=0,last=9]{lcg}

\begin{document}
\title{Accelerometer-based Bed Occupancy Detection for Automatic, Non-invasive Long-term Cough Monitoring}

\name{Madhurananda Pahar$^1$, Igor Miranda$^2$, Andreas Diacon$^3$ and Thomas Niesler$^1$}
\address{
	$^1$Department of Electrical and Electronic Engineering, Stellenbosch University, South Africa \\
	$^2$Federal University of Recôncavo da Bahia, Brazil \\
	$^3$TASK Applied Science, Cape Town, South Africa }
\email{$^1$\{mpahar, trn\}@sun.ac.za, $^2$igordantas@ufrb.edu.br, $^3$ahd@sun.ac.za}


\maketitle

\begin{abstract}
	We present a new machine learning based bed-occupancy detection system that uses the accelerometer signal captured by a bed-attached consumer smartphone.
	Automatic bed-occupancy detection is necessary for automatic long-term cough monitoring, since the time which the monitored patient occupies the bed is required to accurately calculate a cough rate.
	Accelerometer measurements are more cost effective and less intrusive than alternatives such as video monitoring or pressure sensors.
	A 249-hour dataset of manually-labelled acceleration signals gathered from seven patients undergoing treatment for tuberculosis (TB) was compiled for experimentation.
	These signals are characterised by brief activity bursts interspersed with long periods of little or no activity, even when the bed is occupied.
	To process them effectively, we propose an architecture consisting of three interconnected components.
	An occupancy-change detector locates instances at which bed occupancy is likely to have changed, an occupancy-interval detector classifies periods between detected occupancy changes and an occupancy-state detector corrects falsely-identified occupancy changes.
	Using long short-term memory (LSTM) networks, this architecture was demonstrated to achieve an AUC of 0.94.
	When integrated into a complete cough monitoring system, the daily cough rate of a patient undergoing TB treatment was determined over a period of 14 days.
	As the colony forming unit (CFU) counts decreased and the time to positivity (TPP) increased, the measured cough rate decreased, indicating effective TB treatment.
	This provides a first indication that automatic cough monitoring based on bed-mounted accelerometer measurements may present a non-invasive, non-intrusive and cost-effective means of monitoring long-term recovery of TB patients.
	
\end{abstract}





\section{Introduction}
\label{sec:introduction}
Coughing is a symptom of many lung diseases including tuberculosis (TB) and COVID-19. 
Long term cough monitoring, over periods ranging from days to months, may allow the progression of such conditions to be tracked in a cost-effective and non-invasive manner~\cite{pinhas2007methods}.
For example, TB is commonly assessed by the analysis of sputum samples.
This is a time-consuming and costly clinical practice that requires the engagement of trained medical personnel as well as specialised laboratory facilities for the analysis~\cite{de2017assessment}. 
However, experimental evidence suggests that TB patients that are responding to treatment also exhibit a reduction in cough frequency~\cite{proano2017dynamics}. 
Cough monitoring would offer the advantages that it requires neither a laboratory nor medical personnel and is cost-effective since it could be implemented on a smartphone or similar device.

Although our proposed system has wider applications, we consider the particular scenario of long-term cough monitoring in the ward environment of a TB clinic.
The objective is to provide an alternative means of monitoring the success of treatment received by the patients in this facility.
Previously, we have developed a system that can accurately detect coughs from the signal obtained from the tri-axial accelerometer on-board a consumer smartphone~\cite{pahar2021deep}.
The smartphone itself is attached to the bed-frame of the patient under treatment.
By not relying on audio signals, as many alternative approaches do, the system sidesteps the privacy concerns that accompany such audio-based classifiers~\cite{ren2010monitoring}.
Furthermore, since this is not a wearable sensor, it is less intrusive and more convenient.
However, since monitoring must take place continuously and over extended periods, it is necessary to know at which times the patient occupied the bed in order to reliably estimate a cough rate~\cite{mohiuddin2019patient}.
In this work, we consider such bed-occupancy detection using the same acceleration signals.
We note that eventually both the cough detector and the bed occupancy detector can be implemented on the same smartphone as a single integrated system.

The remainder of this paper is structured as follows.
First,  we provide some background on bed occupancy detection and the use of accelerometers for human activity monitoring in Section \ref{sec:background}. 
Next, we describe the compilation of the dataset we use to train and evaluate our algorithms in Section~\ref{sec:data} and the features extracted from this data in Section~\ref{sec:feat-extract}.
The accelerometer-based bed occupancy detection is presented in Section~\ref{sec:bedoccupancydetection} along with the classification strategy in Section~\ref{sec:classification}.
The subsequent application to the monitoring of long-term cough rates is presented in Section~\ref{sec:long-term}. 
Experimental results are shown in Section~\ref{sec:results} and discussed in Section~\ref{sec:discussion}. 
Finally, Section~\ref{sec:conclusion} concludes our paper. 


\section{Background}
\label{sec:background}


\subsection{Bed occupancy detection}


One approach to the detection of bed occupancy is by means of automated video analysis \cite{cournan2016improving}. 
However, video surveillance is intrusive, raises privacy concerns~\cite{sanders1996safety} and our own experience has shown strong resistance from patients to this type of monitoring. 

Another common way of determining bed occupancy is by means of pressure sensors placed under the mattress \cite{jones2006identifying}. 
While this is a very direct way of establishing bed occupancy, it requires specialised and costly equipment.
Furthermore, such pressure sensors have sometimes been found to be sensitive to factors such as the type of mattress and the weight of the patient, leading to incorrect measurements~\cite{taylor2013bed}. 

To our knowledge, bed-occupancy detection based on accelerometer signals has not been reported in the literature before.

\subsection{Accelerometer-based patient monitoring}

Accelerometers are well established as wearable sensors for human physical activity~\cite{montoye1983estimation,treuth2004defining,matthew2005calibration,trost2005conducting,ward2005accelerometer,trost2011comparison}.
For example, accelerometer signals have been used to successfully classify human movement such as walking, running, sitting, standing and climbing the stairs with sensitivities and specificities above 95\%~\cite{bouten1994assessment,mathie2004classification}, even at sampling rates as low as 5~Hz~\cite{randell2000context}.
Similar success has been achieved when using accelerometer signals to distinguish between different walking styles~\cite{ravi2005activity, gafurov2006biometric} and for human fall detection~\cite{bourke2007evaluation}.


Recently, wearable consumer devices such as smartphones with on-board tri-axial accelerometers have been used to classify human activity~\cite{bao2004activity, brezmes2009activity, casale2011human, kwapisz2011activity, bayat2014study, siirtola2012recognizing} and vehicular motion~ \cite{hemminki2013accelerometer}.
While initial studies used simple classifiers such as logistic regression (LR) and multi-layer perceptrons (MLPs), more recently deep neural networks (DNNs) such as convolutional neural networks (CNNs) have offered better performance in recognising human activities from the wearable sensor data~\cite{zhang2015recognizing,lee2017human}.
Among deep approaches, CNNs have been shown to be well-suited to real-time human activity recognition using accelerometer measurements by offering the computational efficiency required by mobile platforms~\cite{ignatov2018real, hassan2018robust}.


\section{Data}
\label{sec:data}

We use a manually annotated dataset of continuous accelerometer measurements obtained from seven patients to train and evaluate the classifiers used to detect bed occupancy. 
All data has been collected as part of this study at a small 24h TB clinic near Cape Town, South Africa.
The clinic accommodates approximately 10 staff and 30 patients in a number of wards each accommodating up to four beds.
Typically, patients spend between 5 and 15 days at the clinic, during which time they undergo treatment and are monitored. 
All patients for whom data was collected were adult male.
No other patient information was gathered due to the ethical constraints of this study.

\subsection{Recording setup}

\begin{figure}
	\centerline{\includegraphics[width=0.5\textwidth]{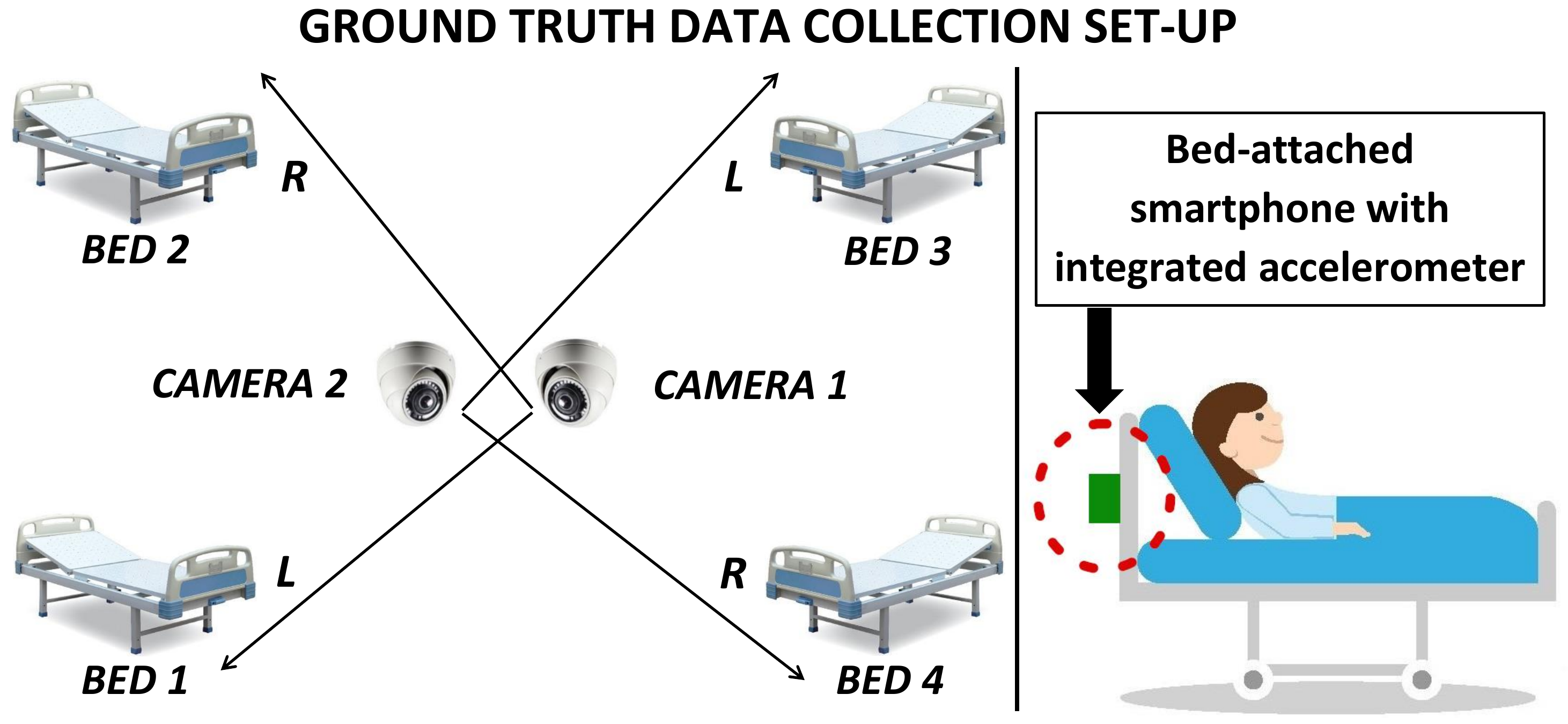}}
	\caption{\textbf{Data collection set-up.} A plastic enclosure housing an inexpensive consumer smartphone (Samsung Galaxy J4) is firmly attached to the back of the headboard of each bed and an Android data recording application continuously monitors the accelerometer signal. To detect the bed occupancy, two ceiling mounted cameras were used. The position of the beds, cameras and camera view-angles are shown. Camera 1 monitors Beds~1 and~2, while Camera~2 monitors Beds~3 and~4. }
	\label{fig:data-collection}
\end{figure}

The data collection process is shown in Figure \ref{fig:data-collection}.
Acceleration signals were captured by the on-board tri-axial accelerometer of a consumer smartphone (Samsung Galaxy J4) that had been firmly attached to the back of the headboard of each of the four beds in a ward.
Data capture software implemented on this device continuously monitored the accelerometer signals at a sampling frequency of $100$ Hz.
To reduce the volume of data captured, a simple energy threshold detector was implemented to exclude long periods with no measured acceleration.
To further reduce the volume of data and also to remove dependence on the orientation of the smartphone, only the vector magnitude of the three tri-axial acceleration components was recorded, as indicated in Equation~\ref{eq:acc_signal}.
Here $a_x(t)$, $a_y(t)$ and $a_z(t)$ are the measured tri-axial accelerations for a particular patient and $t$ is the time from the start of monitoring.
The total period of monitoring varied between 19 and 65 hours for the seven patients in our dataset.
Finally, 
the respective acceleration signal $a(t)$ was normalised so that its maximum value corresponded to a value of 1 after completion of the recording for a patient. 

\begin{eqnarray}
	a(t) &=& \sqrt{ a^2_x(t) + a^2_y(t) + a^2_z(t)} 
	\label{eq:acc_signal}
\end{eqnarray}

In addition to the acceleration signals captured by the smartphone, continuous simultaneous video recordings were made using two ceiling-mounted cameras, as shown in Figure~\ref{fig:data-collection}.
These video recordings were used only for the manual annotation of the captured acceleration signals and hence to provide accurate ground truth labels in terms of when each monitored bed was occupied.


\subsection{Data Annotation}
\label{subsec:annot}

\begin{figure}
	\centerline{\includegraphics[width=0.5\textwidth]{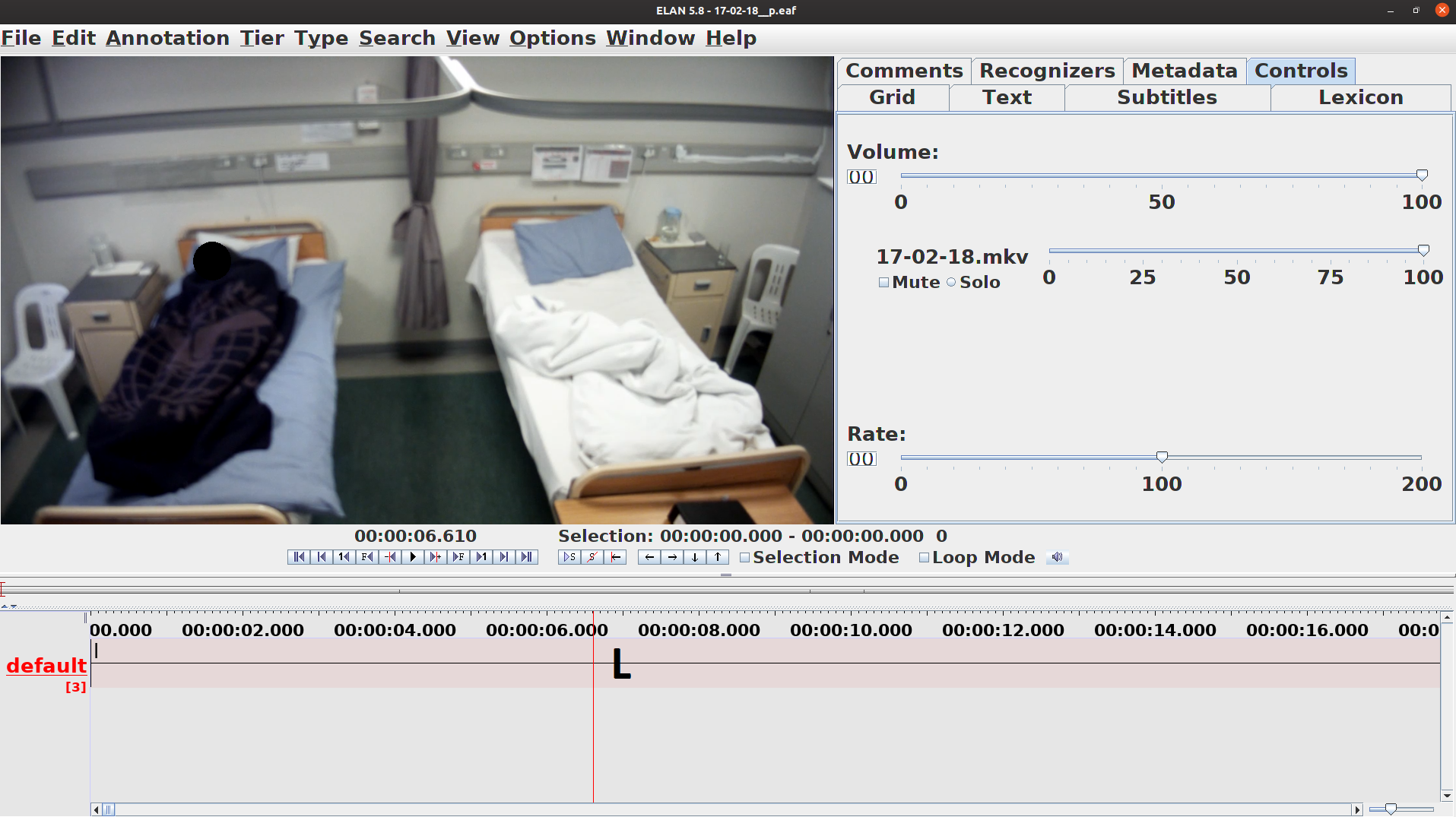}}
	\caption{\textbf{Annotation in ELAN.} The bed on the left is occupied, hence the annotation label `L'. }
	\label{fig:annot_process}
\end{figure}

The accelerometer magnitude signals $a(t)$ were annotated using the ELAN multimedia software (shown in Figure~\ref{fig:annot_process}), since this allowed consolidation with the video data from the ceiling-mounted cameras~\cite{wittenburg2006elan}.
In this way the periods during which each bed was occupied could be accurately labelled, providing ground truth for our experiments. 
As every camera can view only two beds, the labels indicate whether the left bed (L), the right bed (R) or both beds (B) are occupied.

For a particular patient and therefore a particular acceleration signal $a(t)$, this ground truth annotation is represented by the signal $n(t)$, shown in Figure \ref{fig:IP-2019-06-27-acc-annot}.

\begin{equation}
	n(t) = \quad
	\begin{cases}
		1 & \text{if bed is occupied at time $t$,}\\
		0 & \text{if bed is not occupied at time $t$.}
	\end{cases}
	\label{eq:annot}
\end{equation}

%

We note further that, for all our patients, the bed is initially empty and hence $n(0) = 0$.
Furthermore, $n(t)$ switches from 0 to 1 at the instant the patient begins to get into the bed, and switches from 1 to 0 at the instant the patient has completely left the bed.
These conventions allowed the accurate determination of these bed occupancy changes from the joint inspection of the accelerometer and the video footage during manual annotation.

\subsection{Compiled dataset}

%
%
%

\begin{figure}
	\centerline{\includegraphics[width=0.5\textwidth]{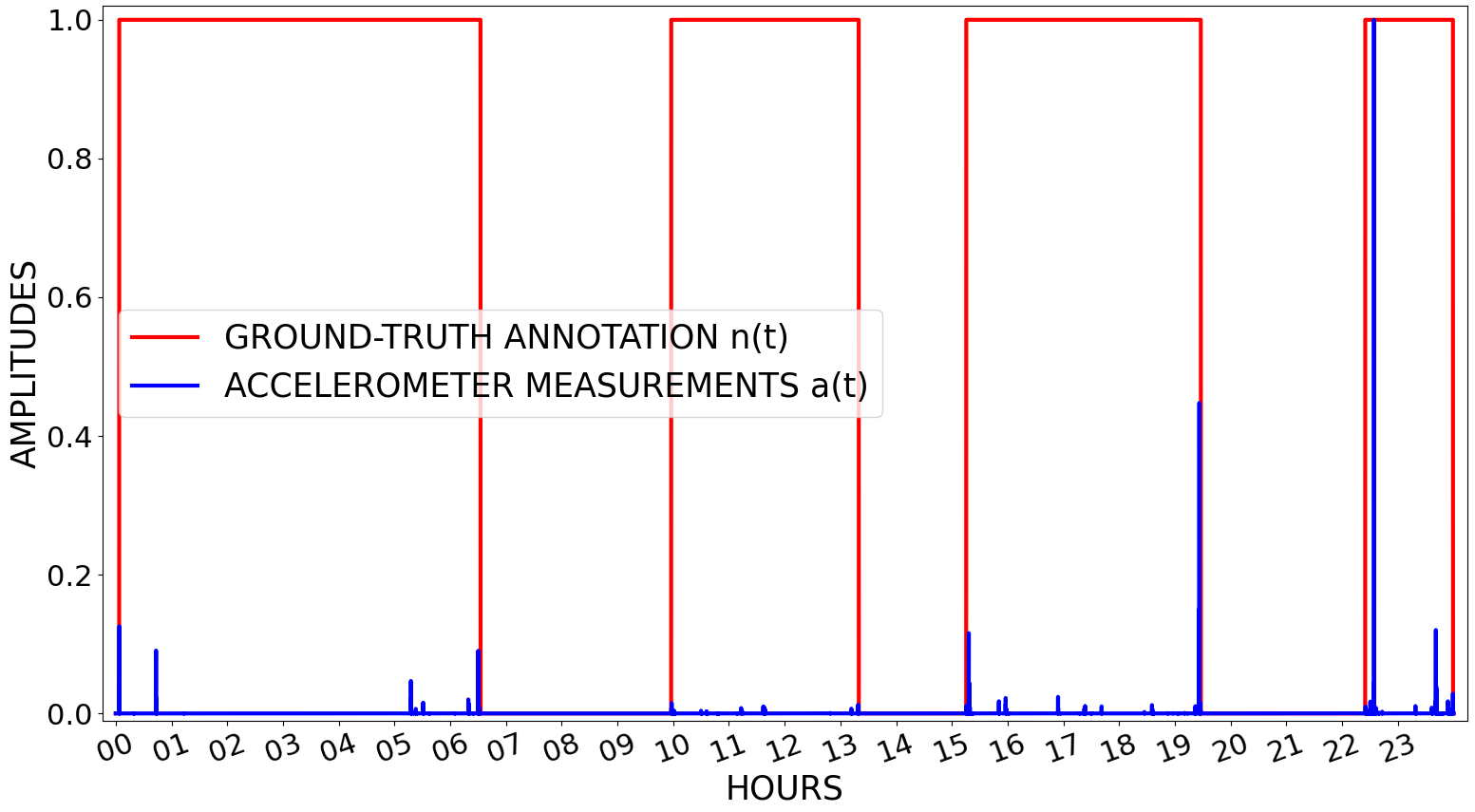}}
	\caption{\textbf{Example accelerometer signal $a(t)$ and ground-truth annotation $n(t)$ over a 24-hour period.} The figure shows that the patient is often not very active while in the bed, indicated by usually very low accelerometer signal amplitudes.  }
	\label{fig:IP-2019-06-27-acc-annot}
\end{figure}

Data was captured from a total of seven patients, each of whom was continuously monitored for a period of between one and three days, as listed in Table~\ref{table:gt_data}. 
In total, 249 hours of acceleration and video data was collected.
All of this data was annotated as described in the previous section.
Table~\ref{table:gt_data} shows that, of the 249 hours of collected data, patients occupied their beds for only 95.5 hours.
Furthermore, a total of only 104 occupancy changes (either getting into or getting out of bed) were identified.
Thus, despite the many hours in our dataset, it remains very sparse in terms of bed occupancy changes.
Table \ref{table:gt_data} summarizes the ground truth dataset. 
A sample accelerometer measurements and annotated data indicating a patient's bed-occupancy pattern are shown in Figure \ref{fig:IP-2019-06-27-acc-annot}. 

\begin{table}[h!]
	\caption{\textbf{Dataset Summary.} Bed numbers and cameras are as indicated in Figure \ref{fig:data-collection}. The number of continuous hours of data that were captured and annotated are indicated, as well as the portion of this period during which the bed was occupied. Occupancy changes refers to the total number of times the patient left or entered the bed in the observation period. } 
	\centering 
	\begin{center}
		\begin{tabular}{ c c c c c c }
			\hline
			\hline
			\multirow{2}{*}{\textbf{Patient}} & \multirow{2}{*}{\textbf{Bed}} & \multirow{2}{*}{\textbf{Camera}} & \textbf{Total} & \textbf{Occupied} & \textbf{Occupancy} \\
			& &  & \textbf{hours} & \textbf{hours} & \textbf{changes} \\
			\hline
			\hline
			1 & 2 & 1 & 65 & 23.33 & 18 \\\hline
			2 & 3 & 2 & 57 & 14.47 & 20 \\ \hline
			3 & 4 & 2 & 45 & 11.67 & 18 \\ \hline
			4 & 4 & 2 & 21 &  9.02 & 10 \\ \hline
			5 & 1 & 1 & 21 & 15.89 & 12 \\ \hline
			6 & 3 & 2 & 19 &  9.42 & 14 \\ \hline
			7 & 2 & 1 & 21 & 11.71 & 12 \\ \hline
			\textbf{Total} &  \textbf{---} & \textbf{---} & \textbf{249} & \textbf{95.51} & \textbf{104} \\
			
			\hline
			\hline
		\end{tabular}
	\end{center}
	\label{table:gt_data}
\end{table}


\section{Feature Extraction} 
\label{sec:feat-extract}

\begin{figure}
	\centerline{\includegraphics[width=0.5\textwidth]{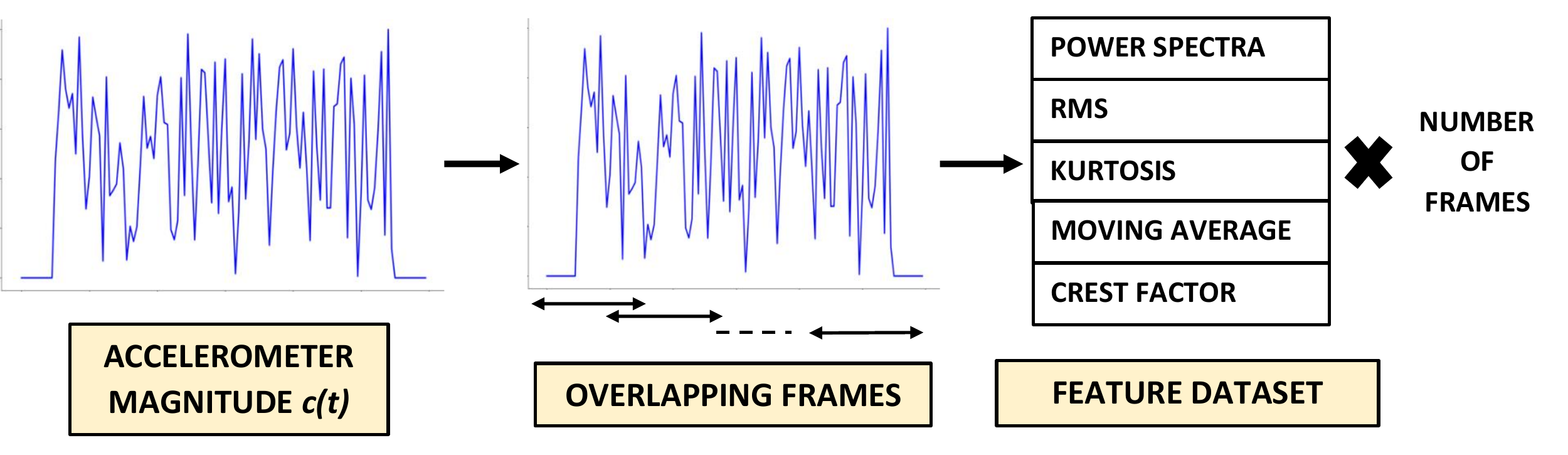}}
	\caption{\textbf{Feature extraction process:}  Power spectra, RMS, kurtosis, moving average and crest factor are extracted from the accelerometer measurements a(t). Number of frames ($C$) is a feature extraction hyperparameter mentioned in Table \ref{table:feat-hyperparameter}.}
	\label{fig:feat-extract}
\end{figure}

The accelerometer magnitude signal $a(t)$ is split into overlapping frames, and features are extracted from each frame.
The frame length ($\Psi$) as well as the number of frames ($C$) are hyperparameters that are optimised by varying the length of frame skip. 
We calculate the frame skips by dividing the number of samples by the number of frames and take the next positive integer. 
Extracted features include the power spectra, root mean square (RMS), moving average (MA), kurtosis and crest factor; 
as they have shown promising results in our previous studies \cite{pahar2021automatic, pahar2021deep}. 
For a frame with $\Psi$ samples, the power spectra has $\frac{\Psi}{2}+1$ coefficients, while RMS, MA, kurtosis and crest factor are scalar.
Hence, a $\frac{\Psi}{2}+5$ dimensional feature vector is extracted from each frame. 


We consider frames with $\Psi = 32$ and $64$ samples, corresponding to 320 and 640 millisecond intervals.
These frames are shorter than those commonly used to extract features from audio for training and evaluating machine learning classifiers~\cite{takahashi2016acoustic}. 
This is because the accelerometer integrated into smartphone have a lower sampling rate (in our case 100 Hz). 
Using longer frames was observed to lead to deteriorated performance, because the acceleration signal can then no longer be assumed to be approximately stationary. 

Finally, we note that classification will generally consider not individual feature vectors, but a sequence of feature vectors extracted from successive frames.
Thus features will be arranged as a feature matrix of size ($C$, $\frac{\Psi}{2}+5$), with the features themselves along one dimension and the frames along the other.


\section{Bed Occupancy Detection Strategy}
\label{sec:bedoccupancydetection}

Bed occupancy detection based on bed-mounted accelerometer signals is challenging, because these signals are characterised by bursts of activity (as the patient either moves or enters or leaves the bed) separated by often very long intervals of inactivity (Figure~\ref{fig:IP-2019-06-27-acc-annot}).
These characteristics make, for example, the direct application of recurrent neural networks ineffective.
In our preliminary experiments we found that even structures designed specifically with long-term memory in mind, such as long short-term memory (LSTM) networks, are not able to learn effectively when presented with signals such as these.

\begin{figure}
	\centerline{\includegraphics[width=0.5\textwidth]{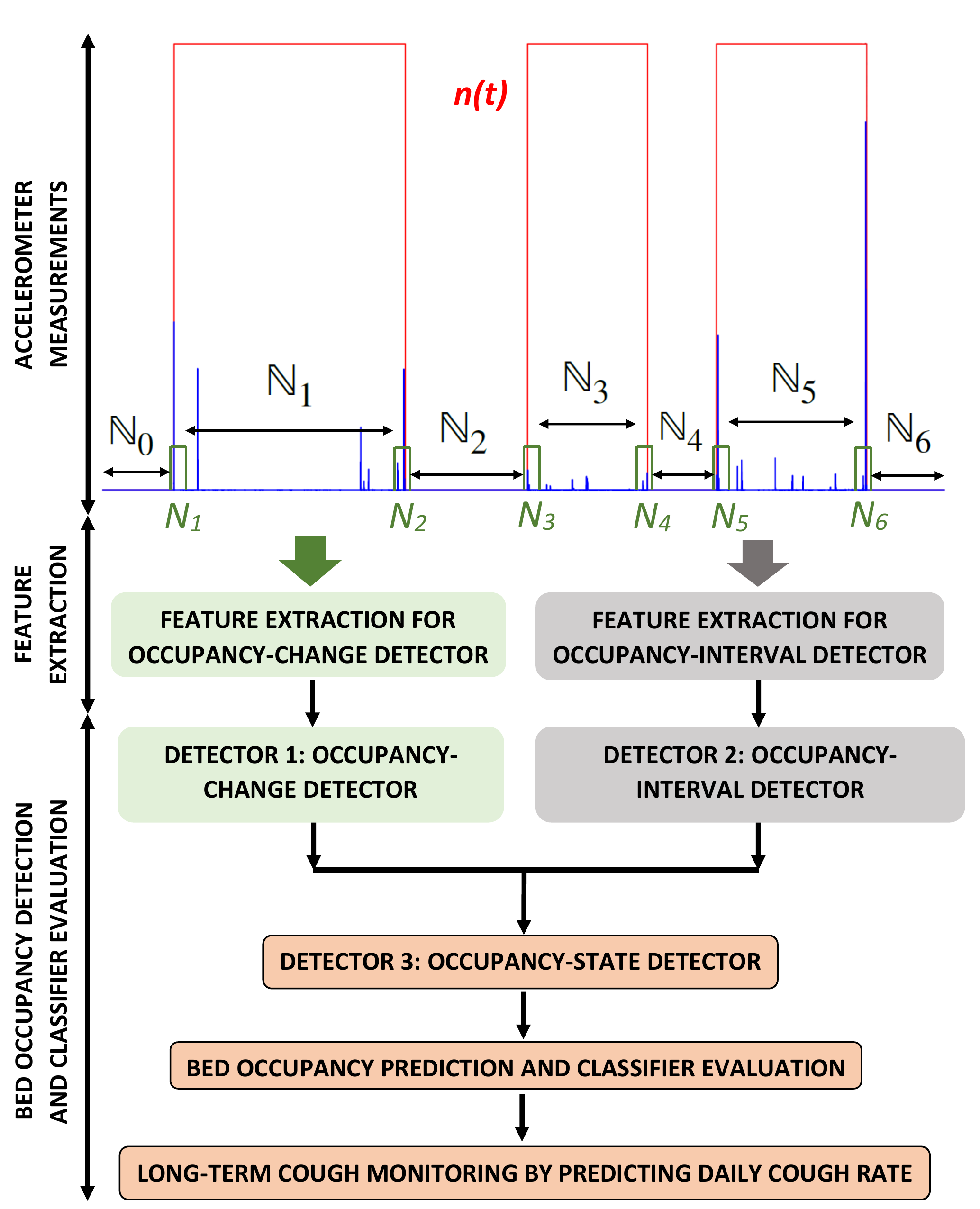}}
	\caption{\textbf{Bed Occupancy Detection Process:} Features are extracted from the accelerometer signals for the occupancy-change and occupancy-interval detection. Using these features, Detector~1 attempts to detect occupancy changes while occupancy intervals are classified by Detector~2. Detector~1 is seen to exhibit low specificity but high sensitivity, while Detector~2 displays high specificity but low sensitivity. Thus, Detector~3, the occupancy-state detector, attempts to correct falsely-detected occupancy changes based on the decisions by Detectors~1 and~2
		, as shown in Figure \ref{fig:predict-process}. In this particular case, there are 6 occupancy changes (the patient went `in' and `out' of the bed 3 times each) i.e. $\mathcal{K} = 6$. So, the number of intervals are: $\mathcal{K}+1 = 7$. 
	}
	\label{fig:class-arc}
\end{figure}

Therefore, we have designed a system that incorporates three interconnected detectors. 
The first is trained to recognise specific portions of the acceleration signal that may be associated with a bed occupancy change, i.e. the short time interval during which a patient enters or leaves the bed.
The second detector is trained specifically to determine whether the interval between two such occupancy changes corresponds to a period during which the bed is occupied or to a period during which it is not occupied.
Finally, the third detector is presented with the outputs of the first two detectors and is designed to correct falsely detected bed occupancy changes.
Figure~\ref{fig:class-arc} illustrates the overall structure of our proposed bed-occupancy detector.

\subsection{Occupancy-Change Detector}\label{subsec:bed-change-feat}

The bed occupancy-change detector (Detector 1) classifies a five-second interval ($T_{oc}=5)$ as either containing an occupancy change or not. 
The acceleration signal $a(t)$ is divided into successive overlapping five-second (500 sample) frames, advancing in steps of 500ms (50 samples).
From each of these five-second frames, we extract $C_{OC}$ sub-frames, each containing $\Psi_{OC}$ samples.
The number of samples by which these sub-frames overlap is determined to ensure that the $C_{OC}$ sub-frames are extracted evenly across the full five-second interval.
For each sub-frame, a feature vector is extracted, and these vectors are arranged as a feature matrix.
Early experimentation showed that a five-second interval exhibits good performance, and this quantity was not optimised further.
However the sub-frame length $\Psi_{OC}$ and sub-frame skip (indirectly determined by $C_{OC}$), both of which influence the dimensionality of this feature matrix, are the hyperparameters that are optimised.

We consider the instant $\tau_k$ at which the $k_{th}$ occupancy change occurs to be the time at which the annotation signal $n(t)$ switches for the $k_{th}$ time, where $k = 1,2,3 \ldots \mathcal{K}$ and where $\mathcal{K} = 104$ is the total number of occupancy changes in our dataset, as listed in Table~\ref{table:gt_data}. 
Since a bed is always empty at $t=0$, $\tau_1$ always indicates a transition from an empty to an occupied bed.
Since in our data the bed is also empty at the end of the recording period, $\mathcal{K}$ is always an even integer.

Inspection of our data confirmed that the location of the instant $\tau_k$ within the five-second interval presented to the classifier should depend on whether the patient is entering or leaving the bed.
When the patient enters the bed, there is almost no activity before  $\tau_k$, while when the patient leaves the bed, there is almost no activity after $\tau_k$.
Hence, during classifier training, the five-second interval $N_k$ surrounding the time instant $\tau_k$ that the classifier is presented with to make its decision is specified differently for these two cases, as shown in Equation~\ref{eq:bed-change}.

\begin{equation}
	N_k = \left\{ \begin{array}{lll} 
		t:& \tau_k - \frac{T_{oc}}{5} \leq t \leq \tau_k + \frac{4T_{oc}}{5} & \textrm{for}\, k \,\textrm{odd} \\[3mm]
		t:& \tau_k - \frac{4T_{oc}}{5} \leq t \leq \tau_k + \frac{T_{oc}}{5} & \textrm{for}\,k \,\textrm{even}
	\end{array}
	\right.
	\label{eq:bed-change}
\end{equation}

Thus, the $N_k$ are the time intervals from 1 sec before to 4 sec after the instants $\tau_k$ at which the bed occupancy changes from unoccupied to occupied and the time intervals from 4 sec before to 1 sec after the instants $\tau_k$ at which the bed occupancy changes from occupied to unoccupied.
We will refer to the $N_k$ as ``occupancy-changes" to distinguish them from the ``occupancy-intervals" considered in the next section.


%

Since most of the acceleration signal is not associated with an occupancy change, the data is highly unbalanced in terms of the two classification classes.
In fact, Table~\ref{table:gt_data} shows that there are only 104 occupancy changes.
For our five-second analysis intervals, this corresponds to a total of only 8.67 minutes of the 249-hour dataset.
Since such imbalance can affect machine learning detrimentally~\cite{van2007experimental,krawczyk2016learning},  we have applied the synthetic minority over-sampling technique (SMOTE) to balance data during training~\cite{chawla2002smote, lemaitre2017imbalanced, windmon2018tussiswatch}.  
This technique oversamples the minor class by generating synthetic samples, as an alternative to for example random oversampling. 
%
%
We have also implemented other extensions of SMOTE such as borderline-SMOTE \cite{BlagusSMOTE, han2005borderline, nguyen2011borderline} and adaptive synthetic sampling \cite{he2008adasyn}. However, the best results were obtained by using SMOTE without any modification. 

At classification time, the measured acceleration magnitude $a(t)$ is presented to the occupancy-change detector, which provides a sequence of hypothesised time instants $\hat{\tau}_k$ and associated intervals $\hat{N}_k$ at which the occupancy of the bed is likely to have changed. 
Early experimental evaluation revealed that this approach allows occupancy-change intervals to be identified with high sensitivity (above 99\%) but low specificity. 
This means that, although most occupancy changes are predicted correctly, other activities, such as movement of the patient while in the bed, have also (wrongly) been classified as occupancy changes.

\subsection{Occupancy-Interval Detector}\label{subsec:bed-interval-feat}

We now consider the time interval between two consecutive occupancy-changes, and focus on the task of determining whether this interval is associated with the bed being occupied or the bed being empty.
We will refer to these as ``occupancy-intervals" ($\mathbb{N}_k$) to distinguish them from the ``occupancy-changes" ($N_k$) considered in the previous section.
The occupancy-intervals $\mathbb{N}_k$ between occupancy-changes $N_{k}$ and $N_{k+1}$ is given by:

\begin{equation}
	\mathbb{N}_k = \left\{ t: N_k < t < N_{k+1}  \right\} 
	\label{eq:occ-interval}
\end{equation}

\noindent where $k = 0, 1, 2, 3 \ldots {\mathcal{K}+1}$ and where $\mathbb{N}_0$ and $N_{\mathcal{K}+1}$ indicate the start and the end of the signal respectively. 

Note that since there are $\mathcal{K}$ occupancy changes, there are $\mathcal{K}+1$ occupancy-intervals.
Furthermore, $\mathbb{N}_0$ indicates the time interval before the first occupancy change and in our data always indicates an initial interval during which the bed is unoccupied, while $\mathbb{N}_\mathcal{K}$ is the interval following the last occupancy change, during which the bed is also empty.






For classification, we divide each occupancy-interval $\mathbb{N}_k$ into ten-second (1000 sample) non-overlapping frames.
From each of these ten-second frames, we extract $C_{OI}$ sub-frames, each containing $\Psi_{OI}$ samples.
The number of samples by which these sub-frames overlap is determined to ensure that the sub-frames are extracted evenly across the full ten-second interval.
For each sub-frame, a feature vector is extracted, and these vectors are arranged as a feature matrix.
Finally, the feature matrices extracted from each ten-second frame in the occupancy-interval $\mathbb{N}_k$ are averaged.
In this way, information about the entire occupancy-interval is encoded as a fixed-dimension feature matrix. 
This method of feature extraction is the result of extensive experimentation.
In particular, approaches such as the direct application of LSTMs to features extracted from $\mathbb{N}_k$ in a more conventional way were, for example, not effective.

According to Table \ref{table:gt_data}, there are total 104 occupancy changes, thus there are 105 occupancy-intervals. 
This means that we generate a dataset containing only 105 feature matrices.
This number is small, especially with a view to training DNNs~\cite{pan2012investigation}. 
Thus, in order to provide justification for deeper architectures, two shallow classifiers (logistic regression and multilayer perceptron) were evaluated in addition to the CNN and LSTM for the occupancy-interval detector.

At classification time, the measured acceleration signal $a(t)$ as well as the hypothesised occupancy-intervals $\hat{\mathbb{N}}_k$, which are derived directly from the hypothesised occupancy-change intervals $\hat{N}_k$ provided by Detector~1, are presented to the occupancy-interval detector, which in turn provides a classification decision $f_{OI}(\hat{\mathbb{N}_k})$ for each occupancy-interval $\hat{\mathbb{N}}_k$ as shown in Equation~\ref{eq:occ-stable-decision}.

\begin{equation}
	f_{OI}\left(\hat{\mathbb{N}}_k \right) = \left\{ \begin{array}{lll}
		1 && \textrm{when occupied} \\[3mm]
		0 && \textrm{when unoccupied}
	\end{array}
	\right. 
	\label{eq:occ-stable-decision}
\end{equation}

Early experimental evaluation showed that this approach allows occupancy-intervals to be identified with high specificity (above 99\%) but lower sensitivity.
Periods during which the bed is unoccupied are reliably identified, but for some patients, who are very quiet when sleeping, an occupied bed was identified as an empty bed. 

The misclassifications by Detector~1 and Detector~2 are addressed by a third and final component, the occupancy-state detector.

\subsection{Occupancy-State Detector}
\label{sec:detector3}

\begin{figure}[!ht]
	\centerline{\includegraphics[width=0.5\textwidth]{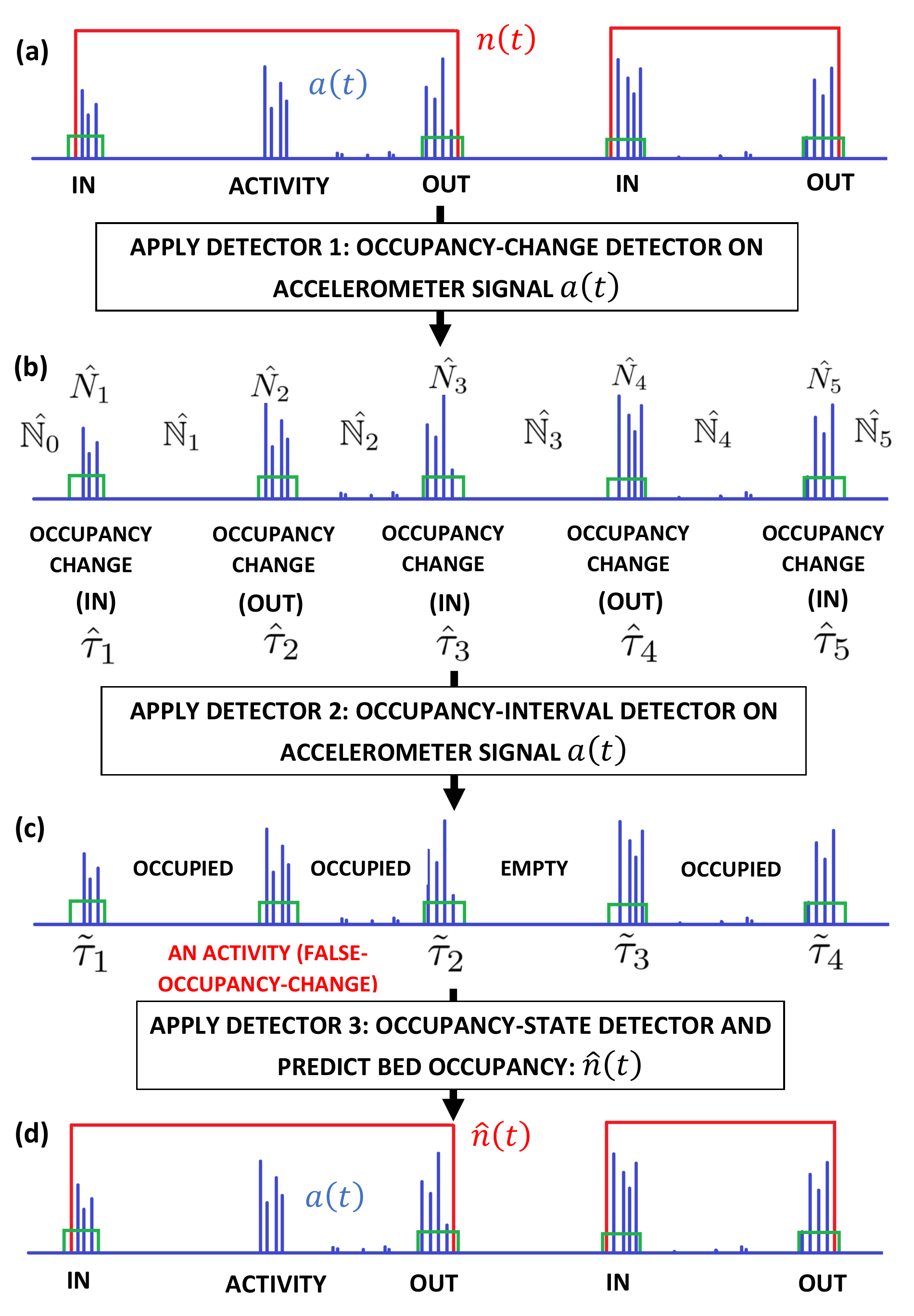}}
	\caption{\textbf{Bed Occupancy Detection Process.} 
		\textbf{(a)} The measured acceleration signal $a(t)$ and its ground-truth annotation $n(t)$.
		\textbf{(b)} The occupancy changes $N_i$ are hypothesised by the occupancy-change detector (Detector~1) and labelled as `in' and `out' successively, while the occupancy-interval detector (Detector~2) decides whether each interval  $\mathbb{N}_i$ between these occupancy changes is classified as `occupied' or `unoccupied'.
		\textbf{(c)} the occupancy-state detector (Detector~3) considers the decisions by Detectors~1 and~2 and reclassifies some occupancy changes asserted by Detector~1 from `change' to `activity', meaning the accelerometer signal was associated with patient movement in the bed and not with the patient entering or leaving the bed.
		\textbf{(d)} This automatically-determined bed occupancy $\hat{n}(t)$ is compared with the hand-annotated bed occupancy $n(t)$ in order to generate the results shown in Table~\ref{table:loo-class-summary-final-bed-presence}.
	}
	\label{fig:predict-process}
\end{figure}

As pointed out in the previous two sections, the occupancy-change detector exhibits high sensitivity but low specificity while the occupancy-interval detector high specificity but low sensitivity.
A third detector, the occupancy-state detector, corrects some of the resulting  errors by considering the outputs of both the occupancy-change and the occupancy-interval detectors.
The process begins by viewing each occupancy-change $\hat{N}_k$ hypothesised by Detector~1 as a true occupancy change, and labelling all occupancy-intervals accordingly.
This results in a labelling sequence  for the occupancy-intervals that alternates at every hypothesised occupancy change and where the first interval $\hat{\mathbb{N}}_0$ is always assumed to be unoccupied. 
Now the hypothesised changes are considered in turn, beginning with the first.
In each case, if the occupancy-intervals to the left and right of a hypothesised change are the same, indicating a false positive by Detector~1, the hypothesised occupancy-change is discounted.
This process is illustrated in Figure~\ref{fig:predict-process}, where a false positive by Detector~1 at $\hat{N}_2$ is removed, thereby associating the accelerometer activity in this interval with patient movement rather than an occupancy change.


Hence, Detector~3 is based on a fixed decision rule.
Experimentation indicated that it is so far not possible to improve on this simple rule by for example making Detector~3 a trainable architecture.
This may be due to the very small training set for this detector, which consists of only 104 hypothesised occupancy changes and 105 occupancy intervals.


\section{Classification Process}
\label{sec:classification}

Classification proceeds as follows.
First, the accelerometer magnitude signal $a(t)$ is divided into successive five-second frames overlapping by 1 second and from which feature matrices are extracted, as described in Section~\ref{subsec:bed-change-feat}.
Detector~1, the occupancy-change detector, classifies each such 5-second frame, and results in a set of time instants $\hat{\tau}_k$ and associated intervals $\hat{N}_k$ at which occupancy changes are likely to have occurred (Figure \ref{fig:predict-process}).
From these intervals $\hat{N}_k$, the corresponding occupancy-intervals $\hat{\mathbb{N}}_k$ are determined, and these are used to extract a new set of feature matrices as described in Section~\ref{subsec:bed-interval-feat}.
These are presented to the occupancy-interval detector, which labels each interval  $\hat{\mathbb{N}}_k$ as either occupied or unoccupied.
Finally, these labels for the $\hat{\mathbb{N}}_k$, together with the intervals $\hat{N}_k$ hypothesised by Detector~1, are processed by Detector~3, resulting in a final set of hypothesised occupancy instants $\tilde{\tau}_k$, which allow a bed-occupancy signal $\hat{n}(t)$ to be determined.
These can be compared with the ground truth annotation signal $n(t)$ in order to determine performance, as described in Section \ref{subsec:eval-process}.

\subsection{Classifier Architectures}
We have considered four classifier types to implement the  bed occupancy detection strategy described in the previous section.
These are logistic regression (LR), multilayer perceptron (MLP), convolutional neural network (CNN) and long short-term memory (LSTM) architectures.
Of these, the first two are shallow architectures which have only been used to classifying occupancy-intervals since in this case the training dataset was especially small.

LR models have been found to outperform other state-of-the-art classifiers such as classification trees, random forests, artificial neural networks and support vector machines in some clinical prediction tasks~\cite{christodoulou2019systematic, botha2018detection, le1992ridge}. 
We have used gradient descent weight regularisation as well as lasso ($l1$ penalty) and ridge ($l2$ penalty) estimators during training~\cite{tsuruoka2009stochastic, yamashita2003interior}.
These regularisation hyperparameters are optimised during cross-validation, as described in Section~\ref{subsec:crossvalidation}.

MLP models \cite{taud2018multilayer} are capable of learning non-linear relationships and have for example been shown to be effective when discriminating influenza coughs from other coughs~\cite{sarangi2016design}. 
MLPs have also been applied to tuberculosis coughs~\cite{tracey2011cough} and to cough detection in general~ \cite{liu2014cough, amoh2015deepcough}. 
During training, we have applied stochastic gradient descent with the inclusion of an $l2$ penalty.
This penalty, along with the number of hidden layers have been considered as hyperparameters.

CNNs are a deep neural network architecture which is popular especially in image classification \cite{krizhevsky2017imagenet}. 
CNNs have successfully solved complex tasks such as cough classification \cite{pahar2020covid, pahar_coding_2020, pahar2021tb, pahar2021deep}, face recognition etc \cite{lawrence1997face}. 
Our CNN architecture \cite{albawi2017understanding, qi2017comparison} consists of $\alpha_1$ 2D convolutional layers with kernel size $\alpha_2$, rectified linear unit (RELU) activation functions, a dropout rate $\alpha_3$ and max-pooling. 
The convolutional layers are followed by a dense layer with $\alpha_4$ units, also using RELU activation functions and then another dense layer with 8 units with rectified linear unit as activation function has been applied. 
Finally, the network is terminated by a two-dimensional softmax layer, indicating an occupancy change for Detector~1 and an occupied bed for Detector~2.
During training, feature matrices are presented to the classifier in batches of $\xi_1$ for a total of $\xi_2$ epochs.


LSTMs are a recurrent neural network (RNN) architecture which has been found to be effective for the modelling of sequential patterns in data.  
Originally suggested in~\cite{hochreiter1997long}, LSTMs have been successfully used in automatic cough detection and classification in our previous studies \cite{miranda2019comparative, pahar2021deep, pahar2020covid}. 
It can also be used for other sort of acoustic feature detection as well \cite{marchi2015non, amoh2016deep}. 
Our LSTM architecture \cite{sherstinsky2020fundamentals} uses $\beta_1$ LSTM units with RELU activation functions and a dropout rate $\alpha_3$. 
The LSTM layer are followed by dense layer with $\alpha_4$ units, also using RELU activation functions and then another dense layer with 8 units and rectified linear unit as activation function has been applied. 
Finally, the network is terminated by a two-dimensional softmax layer, indicating an occupancy change for Detector~1 and an occupied bed for Detector~2.
During training, feature matrices are presented to the classifier in batches of $\xi_1$ for a total of $\xi_2$ epochs, where the feature vectors comprising each feature matrix are presented to the RNN in sequence.


The hyperparameters associated with these classifiers, as listed  in Table~\ref{table:class-hyperparameter}, were optimised using a leave-one-out cross-validation scheme.

\subsection{Training and Hyperparameter Optimisation}
\label{subsec:crossvalidation}


\begin{table*}[h]
	\caption{\textbf{Feature extraction hyperparameters} used in feature extraction for occupancy change and occupancy interval detection. Frame lengths are varied between 320 and 640 milliseconds and the number of frames is varied between 20 and 100. } 
	\centering 
	\begin{center}
		\begin{tabular}{ c | c | c }
			\hline
			\hline
			\textbf{Hyperparameter} & \textbf{Description} & \textbf{Range} \\
			\hline
			
			\hline
			
			$\Psi_{OC}$ & Number of samples per frame used by occupancy-change detector (Detector~1)     & $2^k$ where $k=5, 6$ \\
			\hline
			$C_{OC}$ & Number of frames in feature matrix for occupancy-change detector (Detector~1)  & 20, 50 \\
			\hline
			$\Psi_{OI}$ & Number of samples per frame used by occupancy-interval detector (Detector~2)     & $2^k$ where $k=5, 6$ \\
			\hline
			$C_{OI}$    & Number of frames in feature matrix for occupancy-interval detector (Detector~2)  & 50, 100 \\
			
			\hline
			\hline
		\end{tabular}
	\end{center}
	\label{table:feat-hyperparameter}
\end{table*}

\begin{table*}[h]
	\caption{\textbf{Classifier hyperparameters}, optimised using the leave-one-out cross-validation (Section~\ref{subsec:crossvalidation})  } 
	\centering 
	\begin{center}
		\begin{tabular}{ c | l | c | l  }
			\hline
			\hline
			\textbf{Hyperparameter} & \textbf{Description} & \textbf{Classifier} & \textbf{Range} \\
			\hline
			\hline
			$\nu_1$     & Regularisation strength of penalty ratios     & LR         & $10^i$ where $i=-7,-6,\ldots,6,7$ $(10^{-7}$ to $10^{7})$ \\
			\hline
			$\nu_2$     & $l1$ penalty ratio           & LR         & 0 to 1 in steps of 0.05 \\
			\hline
			$\nu_3$     & $l2$ penalty ratio           & LR         & 0 to 1 in steps of 0.05 \\
			\hline
			$\eta_1$      & No. of hidden layers         & MLP        & 10 to 100 in steps of 10 \\
			\hline
			$\eta_{2}$ & $l2$ penalty ratio           & MLP        & $10^i$ where $i=-7,-6,\ldots,6,7$ $(10^{-7}$ to $10^{7})$ \\
			\hline
			$\eta_{3}$ & Stochastic gradient descent  & MLP        & 0 to 1 in steps of 0.05 \\
			\hline
			$\xi_1$     & Batch Size                   & CNN, LSTM  & $2^k$ where $k=6, 7, 8$\\
			\hline
			$\xi_2$     & No. of epochs                & CNN, LSTM  & 10 to 200 in steps of 20 \\
			\hline
			$\alpha_1$  & No. of Conv filters          & CNN        & $3 \times 2^k$ where $k=3, 4, 5$ \\
			\hline
			$\alpha_2$  & Kernel size               & CNN        & 2 and 3 \\
			\hline
			$\alpha_3$  & Dropout rate               & CNN, LSTM  & 0.1 to 0.5 in steps of 0.2 \\
			\hline
			$\alpha_4$  & Dense layer size             & CNN, LSTM  & $2^k$ where $k=4, 5$ \\
			\hline
			$\beta_1$   & LSTM units                   & LSTM       & $2^k$ where $k=6, 7, 8$ \\
			\hline
			$\beta_2$   & Learning rate                & LSTM       & $10^k$ where $k=-2,-3,-4$ \\ 
			\hline
			\hline
		\end{tabular}
	\end{center}
	\label{table:class-hyperparameter}
\end{table*}

As our dataset contains only seven patients (Table \ref{table:gt_data}), a leave-one-patient-out cross-validation scheme \cite{Sammut2010, wong2015performance} has been used to train and evaluate all classifiers. 
In this scheme, one patient is held out as a test-patient in an outer loop. 
Among the remaining six patients, five are used in an inner loop to train the classifier while the sixth is used as a development set to optimise the hyperparameters listed in Table~\ref{table:class-hyperparameter}. 
The optimised hyperparameters are used to train the classifier using all six patients in the outer loop which is then evaluated on the seventh patient, which had been set aside for independent testing.
This procedure is repeated so that all seven patients are used as as an independent test set in turn.
The area under receiver operating characteristic (AUC) is used to optimise the hyperparameters.
Final classifier performance is determined as the average sensitivity, specificity, accuracy and AUC over the seven outer-loop test sets.

The combination of hyperparameters for which the highest AUC has been obtained inside an outer loop, is defined as the `best hyperparameters' in Table \ref{table:loo-class-summary-bed-change} and \ref{table:loo-class-summary-bed-presence}.

\subsection{Evaluation Process}\label{subsec:eval-process}

%
%
%

We evaluate the performance achieved by our bed-occupancy detection system by comparing the true bed occupancy $n(t)$ with the predicted bed occupancy $\hat{n}(t)$ on a per-sample basis (Figure \ref{fig:predict-process}). 
The actual bed occupancy has been confirmed by manual annotation, as described in Section~\ref{subsec:annot}, while $\hat{n}(t)$ is determined as described at the start of this section.
Since the dataset is imbalanced in terms of the two classes (bed occupied and bed unoccupied), we evaluate classifier performance primarily in terms of the area under the curve (AUC) of the receiver operating characteristic (ROC).
However, we also specify specificity, sensitivity and accuracy in our results.

In addition to evaluating the overall performance of the bed-occupancy detection system, the individual performance of the occupancy-change detector (Detector~1) and the occupancy-interval detector (Detector~2) are also evaluated separately using the same method and performance indicators. 


\section{Long-term Cough Monitoring}
\label{sec:long-term}

We now describe how the bed occupancy detection system (developed in this study) and the cough detection system (developed in our previous work \cite{pahar2021deep}) can be integrated to allow the long-term cough monitoring of a patient who was undergoing TB treatment over a period of 14 days.
The bed-mounted accelerometer signal was recorded for this patient, who was not part of the dataset compiled in Section~\ref{sec:data}.
Furthermore, since the patient in question was undergoing treatment, the normal laboratory analyses used to assess state of health were available. 

\subsection{Cough Counting}\label{subsec:events}

\begin{figure}
	\centerline{\includegraphics[width=0.5\textwidth]{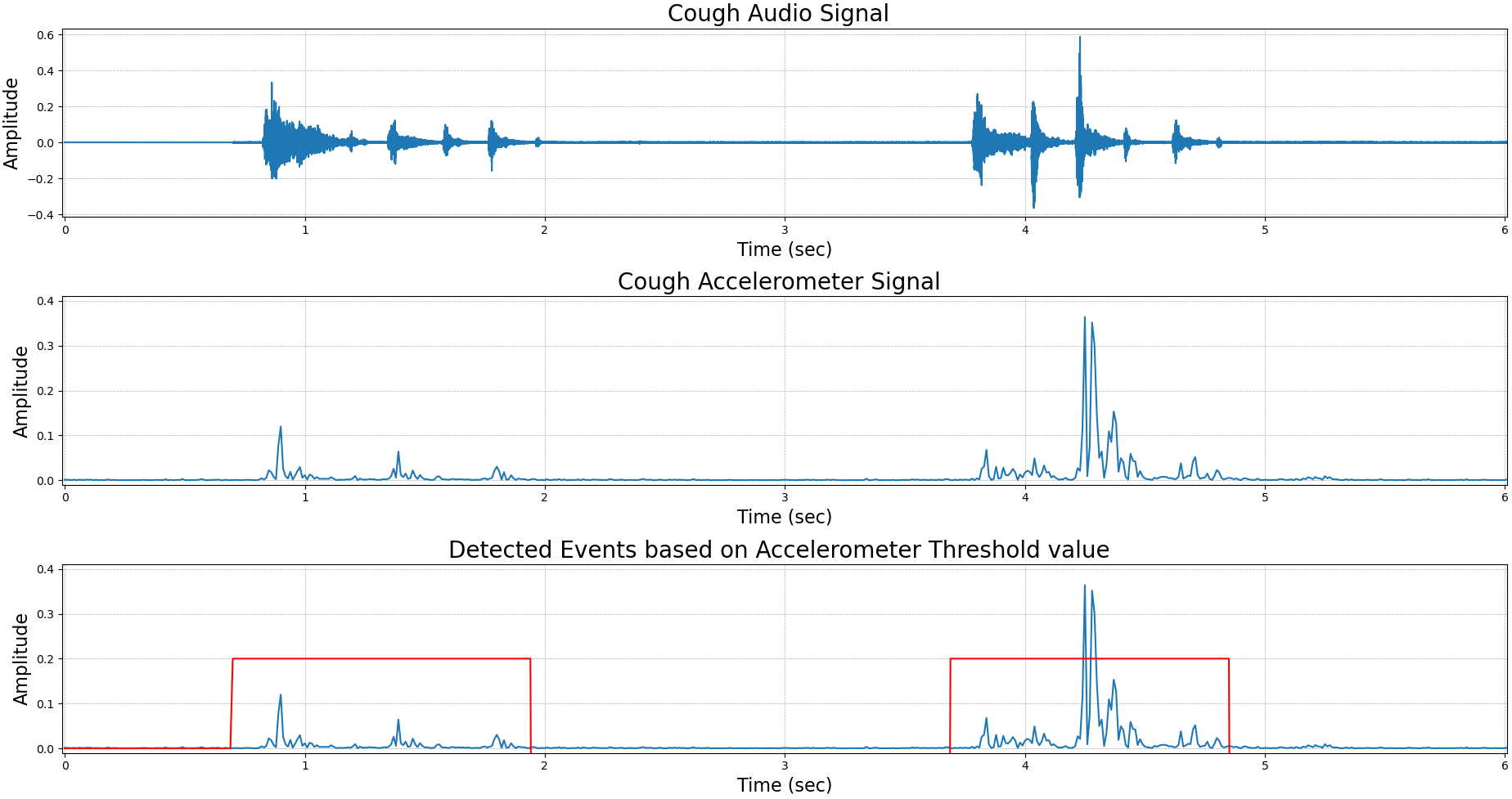}}
	\caption{\textbf{Threshold based event detection.} 
		A simple energy threshold detector is used to isolate portions of the accelerometer signal that are passed to the occupancy-change detector (Detector~1).
	}
	\label{fig:acc-audio-final}
\end{figure}

\begin{figure}[h!]
	\centerline{\includegraphics[width=0.5\textwidth]{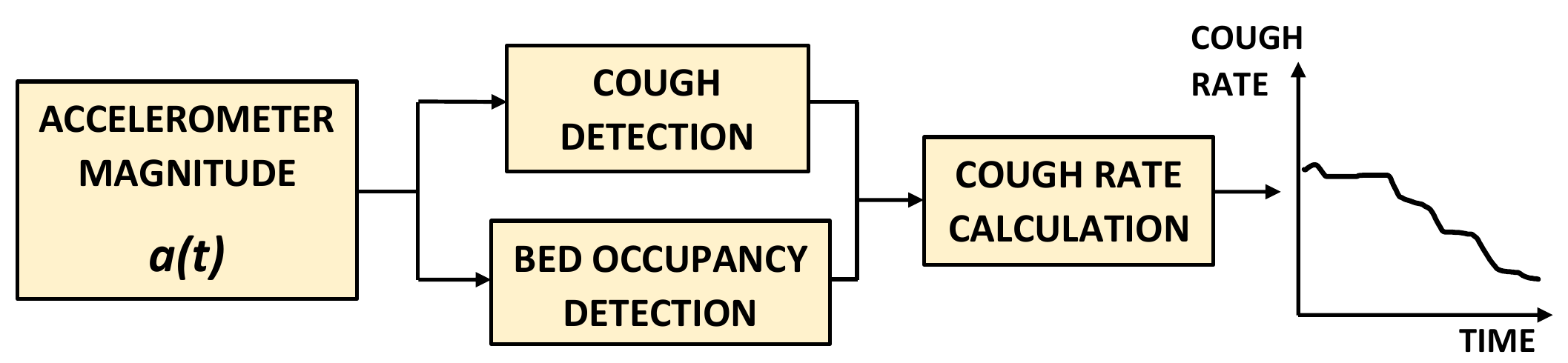}}
	\caption{\textbf{Long-term cough monitoring.} A cough detection system provided the start and end times of all detected coughs in the accelerometer magnitude signal $a(t)$. The bed-occupancy detection system provides the start and end times of all intervals during which the bed is believed to have been occupied. These two sources of information can be used to calculate a cough rate, which is the number of coughs per unit time, for a patient on a certain bed, shown in Figure \ref{fig:daily-cough-predict}. }
	\label{fig:long-term-cough-monitor}
\end{figure}

We have previously developed a system that is able to reliably detect coughs from the same accelerometer signals $a(t)$ we are using for bed-occupancy detection in this work~\cite{pahar2021deep}. 
Since the cough detector uses the accelerometer signal, it is insensitive to the coughs of other patients or visitors.
This is especially useful in a multi-bed ward environment such as the TB clinic at which we are attempting to accomplish automatic long-term cough monitoring. 
A simple threshold-based acoustic event detector, shown in Figure~\ref{fig:acc-audio-final}, was used was used to select the portions of the acceleration signal to pass to the cough detector.
This algorithm extracted sections of the accelerometer signal for which the mean sample amplitude exceeded a small threshold (1\% of the full-scale amplitude) for more than 0.5 seconds.
Figure~\ref{fig:long-term-cough-monitor} provides a high-level diagram of the long-term monitoring system. 

\subsection{Daily Cough Rate}

The average daily cough rate (coughs per 24 hour period) is a means of quantifying how much a patient coughs.
It has been postulated that this figure is related to the state of the patient's heath, and therefore can be used as a means of monitoring \cite{proano2017dynamics}.

\begin{equation}
	\mathbb{R} = \mathbb{C} \times \frac{\mathbb{B}}{24} 
	\label{eq:daily-cough-rate}
\end{equation}

Let $\mathbb{C}$ be the number of coughs detected by the cough detector over a 24-hour period and $\mathbb{B}$ the total time (hours) within this same 24h period that the bed-occupancy detection system believes the patient to have been present in the bed. 
Then, the daily cough rate ($\mathbb{R}$) is determined using Equation~\ref{eq:daily-cough-rate}.

\subsection{Laboratory Indicators}

For the patient whose cough rate we monitored, we also obtained the colony forming unit (CFU) and time to positivity (TTP) values for the same 14-day period.
These are both indicators routinely used to monitor the effectiveness of TB treatment.

The colony forming unit (CFU) count is the number of TB bacterial colonies formed at a certain dilution and it is calculated by Equation~\ref{eq:cfu}. 

\begin{equation}
	\text{CFU} = \log_{10} { \frac{\mathcal{P}_1 + \mathcal{P}_2}{2}  } \times 2 \times 5 \times {10}^{\mathcal{D}}
	\label{eq:cfu}
\end{equation}


The quantities $\mathcal{P}_1$ and $\mathcal{P}_2$ are the number of formed TB colonies in every 1 ml for the two plates used during culturing and $\mathcal{D}$ is the dilution strength, measured in ml \cite{diacon2012time}. 


The time to positivity (TTP) is the number of hours taken for the sputum samples to show signs to being TB positive.
When two plates are cultured, the TPP is calculated as the average: 
\begin{equation}
	\text{TTP} = { \frac{\mathcal{H}_1 + \mathcal{H}_2}{2}  }
	\label{eq:ttp}
\end{equation}
where $\mathcal{H}_1$ and $\mathcal{H}_2$ are the number of hours taken for TB samples to become positive for the two plates respectively~\cite{diacon2012time}. 
Generally, a decrease in CFU is reflected as an increase in TPP, and therefore both can be used as a measure of successful TB treatment~\cite{diacon2012time}. 

The estimated cost of calculating CFUs is around USD 100 and TTPs is around USD 90 and requires usually 3 to 4 weeks per patient to receive the results. 


\section{Results}
\label{sec:results}

In the following, Sections~\ref{sec:results_det1},~\ref{sec:results_det2} and~\ref{sec:results_det3} will present experimental results for Detectors~1,~2 and~3 respectively for the dataset described in Section~\ref{sec:data}.
Then, Section~\ref{sec:results_long} presents results for the long-term cough monitoring described in Section~\ref{sec:long-term}.

\subsection{Detector 1: Occupancy-Change Detection Results}
\label{sec:results_det1}

For the occupancy-change classifier (Detector~1), only the two DNN architectures (CNN and LSTM) were considered.
Both alternatives were trained and evaluated on the data introduced in Section~\ref{sec:data} using the nested cross-validation procedure described in Section~\ref{subsec:crossvalidation}.
The two best-performing (in terms of AUC) systems for each architecture are presented in Table~\ref{table:loo-class-summary-bed-change}.
We note again that the hyperparameters listed in this table were optimised as part of the nested cross-validation, and that the classification performance indicators specificity, sensitivity, accuracy and AUC are averages over the seven outer loops of this process.
In addition, the standard deviation of the AUC, also calculated over the outer loops, is presented and provides and indication of the robustness of the classifiers to variations in the training and testing data.

We see that best bed occupancy change classification performance is achieved by the LSTM when features are extracted using a frame length $\Phi_{OC} = 64$ samples (640ms) and extracting $C_{OC} = 20$ sub-frames from each five-second classification frame.
This system achieves a mean specificity of 71\%, a mean sensitivity of 99\%, a mean accuracy of 85\% and a mean AUC of 0.87. 
We also see, as already commented in Section~\ref{sec:bedoccupancydetection}, that all four systems in Table~\ref{table:loo-class-summary-bed-change} exhibit a high sensitivity, meaning that very few occupancy-changes are missed, but a lower specificity,  meaning that activities such as movement by the patient while in bed are sometimes miss-classified as occupancy changes. 
Nevertheless, we note that the overall success of Detector~1 in identifying occupancy changes implies that the accelerometer signal for this type of event carries some distinguishing patterns that can be used for automatic classification.

\begin{table*}[h]
	\caption{\textbf{Leave-one-patient-out cross-validation results in detecting occupancy-change:} The best two performances are shown for each classifier along with the best hyperparameters. The results show high sensitivity, but low specificity. } 
	\centering 
	\begin{center}
		\begin{tabular}{ c| c c c c c c c c }
			\hline
			\hline
			\multirow{2}{*}{\textbf{Classifier}} & \textbf{Frame} & \textbf{Seg} & \textbf{Mean} & \textbf{Mean} & \textbf{Mean} & \textbf{Mean} & \textbf{SD} & \textbf{Best} \\
			& \textbf{($\Psi_{OC}$)} & \textbf{($C_{OC}$)} & \textbf{Spec} & \textbf{Sens} & \textbf{Acc} & \textbf{AUC} & \textbf{AUC} & \textbf{Hyperparameters} \\
			\hline
			
			\multirow{2}{*}{CNN} & 32 & 20 & 69\% & 99\% & 84\% & 0.86 & 0.0172 & $\xi_1=2^7, \xi_2=100, \alpha_1=24, \alpha_2=2, \alpha_3=0.3, \alpha_4=2^4$  \\
			\cline{2-9}
			& 64 & 50 & 68\% & 99\% & 83.5\% & 0.85 & 0.0278 & $\xi_1=2^7, \xi_2=160, \alpha_1=48, \alpha_2=2, \alpha_3=0.5, \alpha_4=2^5$ \\
			
			\hline
			\multirow{2}{*}{\textit{LSTM}} & \textit{64} & \textit{20} & \textit{71\%} & \textit{99\%} & \textit{85\%} & \textit{0.87} & \textit{0.0179} & \textit{$\xi_1=2^8, \xi_2=180, \alpha_3=0.3, \alpha_4=16, \beta_1=2^7, \beta_2=10^{-3}$} \\
			\cline{2-9}
			& 32 & 50 & 69\% & 99\% & 84\% & 0.86 & 0.0146 & $\xi_1=2^7, \xi_2=140, \alpha_3=0.3, \alpha_4=16, \beta_1=2^7, \beta_2=10^{-2}$ \\
			
			\hline
			\hline
		\end{tabular}
	\end{center}
	\label{table:loo-class-summary-bed-change}
\end{table*}


\subsection{Detector 2: Occupancy-Interval Detection Results}
\label{sec:results_det2}

For the occupancy-interval classifier (Detector~2), two shallow (LR and MLP) and two DNN architectures (CNN and LSTM) were considered.
All four were trained and evaluated on the data introduced in Section~\ref{sec:data} using the nested cross-validation procedure described in Section~\ref{subsec:crossvalidation}.
The best-performing (in terms of AUC) two systems for each architecture are presented in Table~\ref{table:loo-class-summary-bed-presence}.
As before, the listed hyperparameters were optimised during cross-validation, and classification performance is indicated by the averages over the seven outer cross-validation loops.

We see that best classification performance is again achieved by an LSTM, in this case extracting features using a frame length of $\Phi_{OI}=64$ samples (640ms) and 50 sub-frames from each ten-second classification frame.
This system achieves a mean specificity of 99\%, a mean sensitivity of 85\%, a mean accuracy of 92\% and a mean AUC of 0.94. 
We also see, as already commented in Section~\ref{sec:bedoccupancydetection}, that all eight systems in Table~\ref{table:loo-class-summary-bed-presence} exhibit a high specificity, meaning that intervals are rarely classified as ``occupied" when in fact the bed was empty, but a lower sensitivity,  meaning that in some cases the bed is classified as unoccupied when in fact it is not because its occupant exhibits very little movement.

\begin{table*}[h]
	\caption{\textbf{Leave-one-patient-out cross-validation results in detecting intervals:} The best two results are shown for each classifier along with the best hyperparameters. The results show high specificity, but low sensitivity. } 
	\centering 
	\begin{center}
		\begin{tabular}{ c | c c c c c c c c }
			\hline
			\hline
			\multirow{2}{*}{\textbf{Classifier}} & \textbf{Frame} & \textbf{Seg} & \textbf{Mean} & \textbf{Mean} & \textbf{Mean} & \textbf{Mean} & \textbf{SD} & \textbf{Best} \\
			& \textbf{($\Psi_{OI}$)} & \textbf{($C_{OI}$)} & \textbf{Spec} & \textbf{Sens} & \textbf{Acc} & \textbf{AUC} & \textbf{AUC} & \textbf{Hyperparameters} \\
			\hline
			
			\multirow{2}{*}{LR} & 32 & 100 & 98\% & 81\% & 89.5\% & 0.91 & 0.0371 & $\nu_1=10^{-4}, \nu_2=0.35, \nu_3=0.45$ \\
			\cline{2-9}
			& 64 & 50 & 98\% & 80\% & 89\% & 0.90 & 0.0295 & $\nu_1=10^{-3}, \nu_2=0.7, \nu_3=0.3$ \\
			
			
			\hline
			\multirow{2}{*}{MLP} & 64 & 100 & 99\% & 82\% & 90.5\% & 0.92 & 0.0309 & $\eta_1=70, \eta_2=10^{-2}, \eta_3=0.55$ \\
			\cline{2-9}
			& 32 & 50 & 98\% & 81.5\% & 90\% & 0.91 & 0.0205 & $\eta_1=50, \eta_2=10^{-5}, \eta_3=0.2$ \\	
			
			\hline
			\multirow{2}{*}{CNN} & 32 & 100 & 99\% & 84\% & 91.5\% & 0.93 & 0.0249 & $\xi_1=2^8, \xi_2=140, \alpha_1=24, \alpha_2=2, \alpha_3=0.3, \alpha_4=16$ \\
			\cline{2-9}
			& 64 & 100 & 99\% & 83\% & 91\% & 0.92 & 0.0239 & $\xi_1=2^7, \xi_2=100, \alpha_1=48, \alpha_2=3, \alpha_3=0.1, \alpha_4=2^4$ \\
			
			\hline
			\multirow{2}{*}{\textit{LSTM}} & \textit{64} & \textit{50} & \textit{99\%} & \textit{85\%} & \textit{92\%} & \textit{0.94} & \textit{0.0288} & \textit{$\xi_1=2^8, \xi_2=180, \alpha_3=0.3, \alpha_4=2^4, \beta_1=2^7, \beta_2=10^{-3}$} \\
			\cline{2-9}
			& 32 & 100 & 99\% & 83\% & 91\% & 0.93 & 0.0190 & $\xi_1=2^8, \xi_2=180, \alpha_3=0.1, \alpha_4=16, \beta_1=2^8, \beta_2=10^{-3}$ \\
			
			\hline
			\hline
		\end{tabular}
	\end{center}
	\label{table:loo-class-summary-bed-presence}
\end{table*}

\subsection{Detector 3: Bed Occupancy Detection}
\label{sec:results_det3}

\begin{figure}
	\centerline{\includegraphics[width=0.5\textwidth]{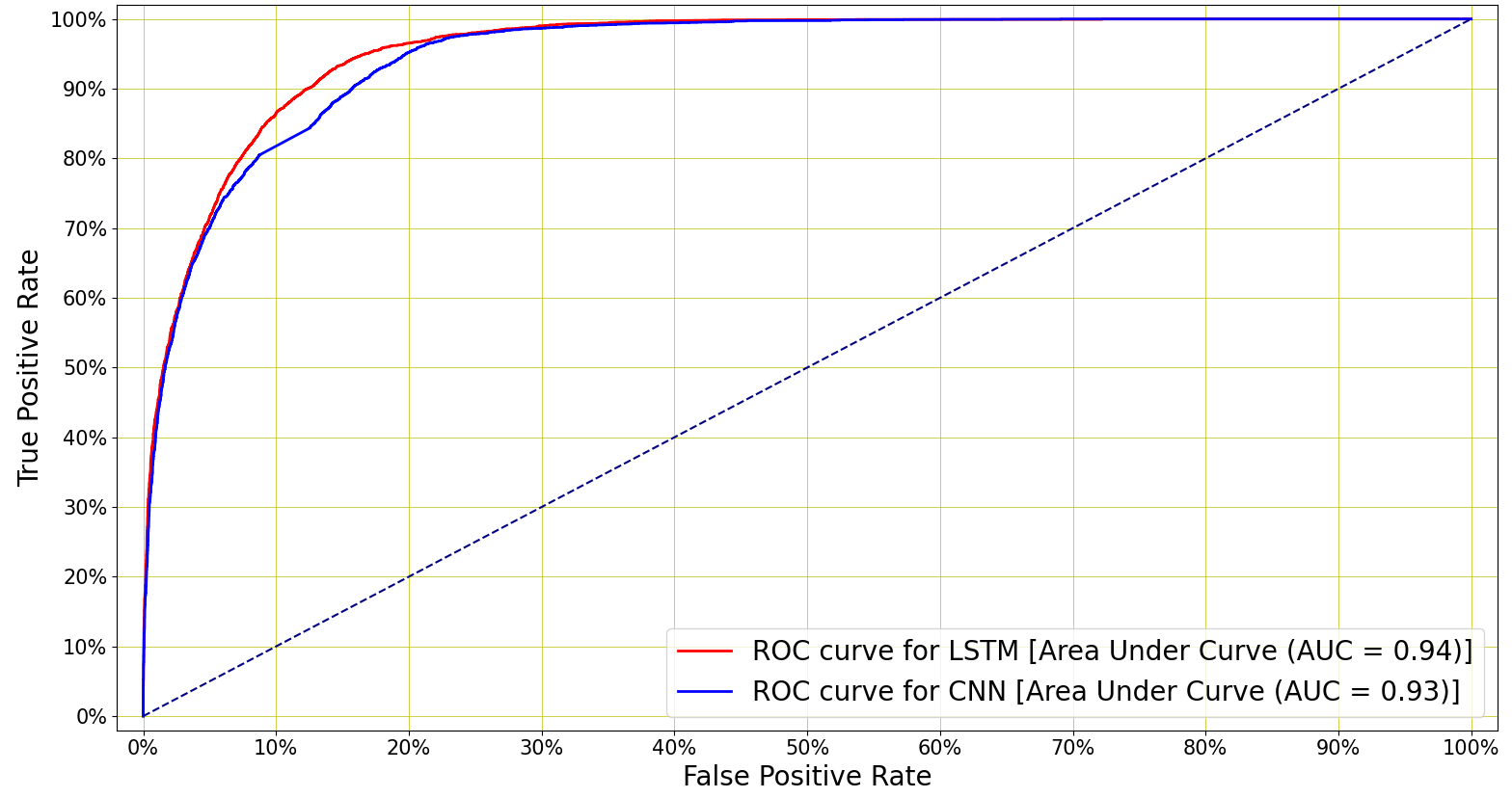}}
	\caption{\textbf{Mean ROC curve for bed occupancy detection}, which shows the best results for LSTM and CNN classifiers. The highest AUC of 0.94 has been achieved for the LSTM classifier, detailed in Table \ref{table:loo-class-summary-final-bed-presence}. }
	\label{fig:mean-ROC}
\end{figure}

As described in Section~\ref{sec:detector3}, the best performing occupancy-change and occupancy-interval classifiers from Sections~\ref{sec:results_det1} and~\ref{sec:results_det2} are considered by the occupancy-state classifier (Detector~3) in order to correct some false positive classification made by Detector~1.
The resulting performance when using either CNN or LSTM classifiers for Detectors~1 and~2 is presented in Table~\ref{table:loo-class-summary-final-bed-presence}, while Figure~\ref{fig:mean-ROC} shows the ROC curves for the two best systems.
The results in this table reflect the overall per-sample classification performance of our accelerometer-based bed-occupancy detection system.
The best performance has been achieved by combining the outputs of the two LSTM classifiers used for Detectors~1 and~2,  resulting in an AUC of 0.94,  a specificity of 91.71\% and a sensitivity of 94.51\%.
Hence the simple procedure implemented by Detector~3 has resulted in an overall system for which both sensitivity and specificity are high.

\begin{table*}[h]
	\centering
	\caption{\textbf{Final leave-one-patient-out cross-validation results in detecting patient's bed occupancy: }The best two results are shown for each classifier along with the best hyperparameters. The results are reliable as they show both high specificity and sensitivity. The highest AUC of 0.94 has been obtained from a LSTM classifier. }
	\begin{tabular}{ c | c c c c c c c c c c }
		\hline
		\hline
		{\multirow{2}{*}{\textbf{Classifier}}} & \multicolumn{2}{c}{\textbf{Detector 1}} &
		\multicolumn{2}{c}{\textbf{Detector 2}} & \textbf{Mean} & \textbf{Mean} & \textbf{Mean} & \textbf{Mean} & \textbf{SD} \\
		
		& \textbf{Frame ($\Psi_{OC}$)} & \textbf{Seg ($C_{OC}$)} & \textbf{Frame ($\Psi_{OI}$)} & \textbf{Seg ($C_{OI}$)} & \textbf{Specificity} & \textbf{Sensitivity} & \textbf{Accuracy} & \textbf{AUC} & \textbf{AUC} \\

		\hline
		\multirow{2}{*}{CNN} & 32 & 50 & 64 & 50 & 93.09\% & 89.77\% & 91.43\% & 0.93 & 0.0205 \\
		\cline{2-10}
		& 64 & 50 & 32 & 100 & 91.01\% & 89.91\% & 90.46\% & 0.92 & 0.0192 \\
		
		\hline
		\multirow{2}{*}{\textit{LSTM}} & \textit{32} & \textit{50} & \textit{64} & \textit{100} & \textit{94.51\%} & \textit{91.71\%} & \textit{93.11\%} & \textit{0.94} & \textit{0.0218} \\
		\cline{2-10}
		& 32 & 20 & 64 & 50 & 92.53\% & 90.88\% & 91.71\% & 0.93 & 0.0271 \\
		
		\hline
		\hline
	\end{tabular}
	\label{table:loo-class-summary-final-bed-presence}
\end{table*}

\subsection{Long-term Cough monitoring}
\label{sec:results_long}

\begin{figure}
	\centerline{\includegraphics[width=0.5\textwidth]{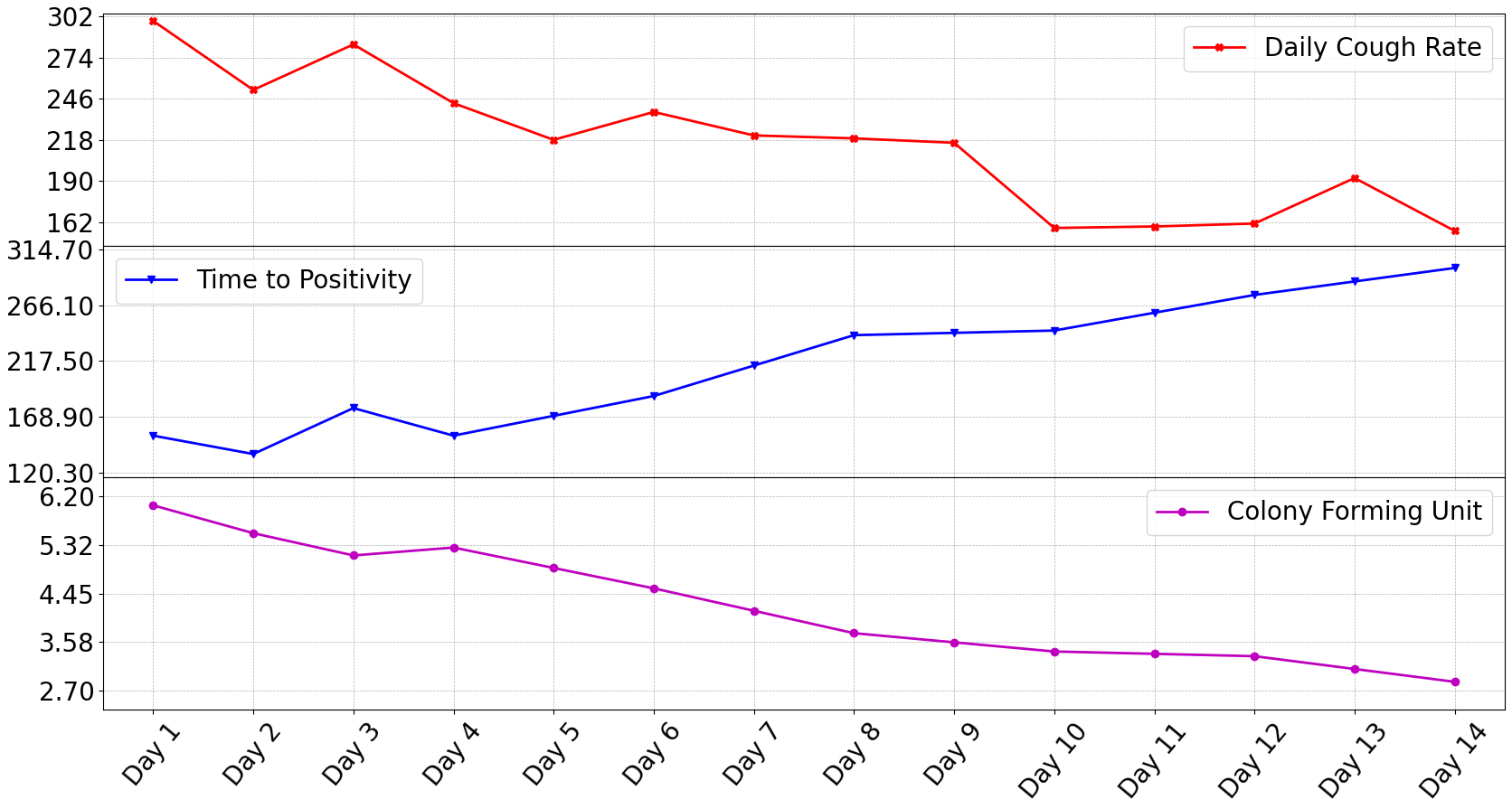}}
	\caption{\textbf{Daily cough rate ($\mathbb{R}$), TTP and CFU} for a patient undergoing treatment for TB over a 14 day period. The graphs indicate that daily cough rate decreases while the TPP increases and CFU decreases over time, suggesting that long-term cough monitoring system presented in this study can be a useful mean of monitoring patient's improvement on TB treatment. }
	\label{fig:daily-cough-predict}
\end{figure}

The best LSTM-based bed-occupancy detection system has been used to determine the daily cough rates $\mathbb{R}$ for a new patient over a 14-day period, as described in Section~\ref{sec:long-term}. 
Figure~\ref{fig:daily-cough-predict} shows the cough rate for this patient, together with the CFU and TPP values over the same period.   
Firstly, we see that the CFU decreases over time, indicating that the number of colonies formed in the sample dilution on average decreases with time.
We also see that the TTP increases, over time, showing that the time taken for a TB sample to become positive is also increasing.
Therefore both microbiological indicators indicate that, in general, TB treatment which the patient was receiving was successful.
Finally, we also see that the daily cough rate decreases over the same time interval, 
This suggests that automatic long-term cough monitoring, such as that implemented by the system we present in this study, may be an alternative viable means of monitoring the health of patients in a TB clinic. 
We note that our observations are also in line with the observations found by \cite{proano2017dynamics},
where the cough frequency was measured manually.


\section{Discussion}
\label{sec:discussion}

The results in Section~\ref{sec:results} demonstrate that bed-occupancy detection is possible to high accuracy based on only the signal captured by a bed-mounted accelerometer.
Since cough detection is also possible using this signal, this allows the cough rate to be accurately determined by allowing periods during which the patient leaves the bed to be discounted from the calculation.
Basing classification decisions only on the accelerometer signal allows a convenient, non-intrusive and privacy-preserving form of cough monitoring.

Since the accelerometer is bed-mounted, the system is insensitive to coughs from persons other than the patient in the bed.
Therefore the system is suited, for example, for a multi-bed ward environment.
The results in Section~\ref{sec:long-term} show that, for a patient undergoing standard TB treatment, the cough rate decreases over a period of 14 days while laboratory indicators such as CFU counts and TPP indicate that the treatment is effective.
Even though we have shown this for only a single patient and therefore further validation is necessary, this is promising empirical evidence to suggest that cough rate can be used as a method of monitoring the effectiveness of treatment for TB patients, and perhaps also patients with other lung ailments.
This could be of benefit because the proposed system is easier, quicker, less invasive and much less costly to implement than laboratory analyses.


\section{Conclusion and Future Work}
\label{sec:conclusion}

We have described a machine learning based bed-occupancy detection system that uses the accelerometer signal captured by a consumer smartphone attached to the patient's bed.
Such bed-occupancy detection is required to allow the implementation of automatic long-term cough monitoring using the same accelerometer signal, since the time which the monitored patient is present in the bed must be known in order to accurately calculate a cough rate.
Using a bed-mounted sensor is more convenient and less intrusive than wearable alternatives or video monitoring, and using only acceleration measurements intrinsically preserves privacy.

For experimental evaluation, we compiled a 249-hour dataset of manually-labelled acceleration signals gathered from seven patients undergoing treatment for tuberculosis (TB).
Inspection of these acceleration signals revealed that they are challenging, since they are characterised by brief activity bursts interspersed with long periods of little or no activity, even when the bed is occupied.
Initial experimentation revealed that recursive neural architectures, such as long short-term memory (LSTM) networks, which in other applications often deliver state-of-the-art performance for tasks that require the modelling of complex sequential data, are ineffective when presented with the acceleration signals in our dataset.
Hence, to process this signal effectively, we developed three interconnected components.
The first, termed the occupancy-change detector, locates instances in time at which the occupancy of the bed is likely to have changed as a result of the patient entering or leaving the bed.
The second, termed the occupancy-interval detector, considers the periods between detected occupancy changes and classifies them as being associated with either an occupied or an unoccupied bed.
The third and final component, termed the occupancy-state detector, uses the results of the first two detectors to correct some of the falsely-identified occupancy changes.

To implement the system, we consider two shallow (liner regression and multilayer perceptron) and two deep (convolutional neural network (CNN) and LSTM) neural architectures.
We employ nested cross-validation to train, optimise and evaluate these architectures and find that a system using LSTM network for both the occupancy-change and the occupancy-interval detectors achieves the best performance, with an area under the ROC curve (AUC) of 0.94.
For all considered combinations, we observed that the occupancy-change detector exhibits a high sensitivity but a lower specificity while the occupancy-interval detector exhibits a high specificity but a lower sensitivity.
Thus, the final occupancy-state detector was able to rectify many falsely-identified occupancy changes and achieve an overall system exhibiting both high sensitivity and high specificity.

As a final step, we implemented a complete cough monitoring system by integrating a previously-developed cough detector, which uses the same acceleration signal, with our proposed bed-occupancy detection system.
The cough detector provides the time instances of detected coughs, while the bed-occupancy detector provides the time intervals during which the bed was occupied.
Together, this information can be used to calculate an accurate estimate of the daily cough rate.
We evaluated this cough rate monitor using acceleration signals gathered over a period of 14 days from a separate patient undergoing TB treatment.
The evolution of the resulting cough rate was compared with the evolution of the colony forming unit (CFU) counts as well as the time to positivity (TPP) determined for sputum samples from the same patient obtained by standard microbiological laboratory analyses.
We were able to show that, as the CFU decreased with time and the TPP increased with time, indicating that TB treatment was effective, the measured cough rate decreased with time.
This provides empirical evidence indicating that cough monitoring based on bed-mounted accelerometer measurements may present a quick, non-invasive, non-intrusive and cost-effective means of monitoring the long-term recovery of TB patients.


As immediate future work,  we aim to apply the presented cough monitoring system to a larger number of patients undergoing tuberculosis treatment, to verify whether the link between cough rate and the clinical indicators remains.
We would also like to determine whether accuracy can be improved by replacing the occupancy-state detector with a trainable architecture.
Although our first attempts at this have not met with success, the extension of our dataset, which is an ongoing process, may make this possible.
Finally, we would like to experiment with other neural architectures, such as residual networks (ResNets~\cite{he2016deep}).

%


\bibliographystyle{IEEEbib}
\bibliography{reference}

\begin{thebibliography}{10}

\bibitem{pinhas2007methods}
Itzhak Pinhas, Avner Halperin, Arkadi Averboukh, Daniel Lange, and Yosef Gross,
\newblock ``Methods and systems for monitoring patients for clinical
  episodes,'' May~24 2007,
\newblock US Patent App. 11/552,872.

\bibitem{de2017assessment}
Gerjo~J de~Knegt, Laura Dickinson, Henry Pertinez, Dimitrios Evangelopoulos,
  Timothy~D McHugh, Irma~AJM Bakker-Woudenberg, Gerry~R Davies, and Jurriaan~EM
  de~Steenwinkel,
\newblock ``Assessment of treatment response by colony forming units, time to
  culture positivity and the molecular bacterial load assay compared in a mouse
  tuberculosis model,''
\newblock {\em Tuberculosis}, vol. 105, pp. 113--118, 2017.

\bibitem{proano2017dynamics}
Alvaro Proa{\~n}o, Marjory~A Bravard, Jos{\'e}~W L{\'o}pez, Gwenyth~O Lee,
  David Bui, Sumona Datta, Germ{\'a}n Comina, Mirko Zimic, Jorge Coronel, Luz
  Caviedes, et~al.,
\newblock ``Dynamics of cough frequency in adults undergoing treatment for
  pulmonary tuberculosis,''
\newblock {\em Clinical Infectious Diseases}, vol. 64, no. 9, pp. 1174--1181,
  2017.

\bibitem{pahar2021deep}
Madhurananda Pahar, Igor Miranda, Andreas Diacon, and Thomas Niesler,
\newblock ``Deep {N}eural {N}etwork based {C}ough {D}etection using
  {B}ed-mounted {A}ccelerometer {M}easurements,''
\newblock in {\em ICASSP 2021 - 2021 IEEE International Conference on
  Acoustics, Speech and Signal Processing (ICASSP)}, 2021, pp. 8002--8006.

\bibitem{ren2010monitoring}
Yonglin Ren, Richard Werner, Nelem Pazzi, and Azzedine Boukerche,
\newblock ``Monitoring patients via a secure and mobile healthcare system,''
\newblock {\em IEEE Wireless Communications}, vol. 17, no. 1, pp. 59--65, 2010.

\bibitem{mohiuddin2019patient}
Abdul~Kader Mohiuddin,
\newblock ``Patient {B}ehavior: an extensive review,''
\newblock {\em Nurse Care Open Access Journal}, vol. 6, no. 3, pp. 76--90,
  2019.

\bibitem{cournan2016improving}
Michele Cournan, Benjamin Fusco-Gessick, and Laura Wright,
\newblock ``Improving patient safety through video monitoring,''
\newblock {\em Rehabilitation Nursing}, 2016.

\bibitem{sanders1996safety}
Pamela~T Sanders, Barbara~J Cysyk, and Mary~A Bare,
\newblock ``Safety in long-term {EEG}/video monitoring,''
\newblock {\em Journal of Neuroscience Nursing}, vol. 28, no. 5, pp. 305--314,
  1996.

\bibitem{jones2006identifying}
M~Howell Jones, Rafik Goubran, and Frank Knoefel,
\newblock ``Identifying {M}ovement {O}nset {T}imes for a {B}ed-{B}ased
  {P}ressure {S}ensor {A}rray,''
\newblock in {\em IEEE International Workshop on Medical Measurement and
  Applications, 2006. MeMea 2006.} IEEE, 2006, pp. 111--114.

\bibitem{taylor2013bed}
Matthew Taylor, Theresa Grant, Frank Knoefel, and Rafik Goubran,
\newblock ``Bed occupancy measurements using under mattress pressure sensors
  for long term monitoring of community-dwelling older adults,''
\newblock in {\em 2013 IEEE International Symposium on Medical Measurements and
  Applications (MeMeA)}. IEEE, 2013, pp. 130--134.

\bibitem{montoye1983estimation}
Henry~J Montoye, Richard Washburn, Stephen Servais, Andrew Ertl, John~G
  Webster, and Francis~J Nagle,
\newblock ``Estimation of energy expenditure by a portable accelerometer.,''
\newblock {\em Medicine and Science in Sports and Exercise}, vol. 15, no. 5,
  pp. 403--407, 1983.

\bibitem{treuth2004defining}
Margarita~S Treuth, Kathryn Schmitz, Diane~J Catellier, Robert~G McMurray,
  David~M Murray, M~Joao Almeida, Scott Going, James~E Norman, and Russell
  Pate,
\newblock ``Defining accelerometer thresholds for activity intensities in
  adolescent girls,''
\newblock {\em Medicine and Science in Sports and Exercise}, vol. 36, no. 7,
  pp. 1259, 2004.

\bibitem{matthew2005calibration}
Charles~E Matthew,
\newblock ``Calibration of accelerometer output for adults.,''
\newblock {\em Medicine and Science in Sports and Exercise}, vol. 37, no. 11
  Suppl, pp. S512--22, 2005.

\bibitem{trost2005conducting}
Stewart~G Trost, Kerry~L Mciver, and Russell~R Pate,
\newblock ``Conducting accelerometer-based activity assessments in field-based
  research,''
\newblock {\em Medicine and Science in Sports and Exercise}, vol. 37, no. 11,
  pp. S531--S543, 2005.

\bibitem{ward2005accelerometer}
Dianne~S Ward, Kelly~R Evenson, Amber Vaughn, A~Brown Rodgers, and Richard~P
  Troiano,
\newblock ``Accelerometer use in physical activity: best practices and research
  recommendations.,''
\newblock {\em Medicine and Science in Sports and Exercise}, vol. 37, no. 11
  Suppl, pp. S582--8, 2005.

\bibitem{trost2011comparison}
Stewart~G Trost, Paul~D Loprinzi, Rebecca Moore, and Karin~A Pfeiffer,
\newblock ``Comparison of accelerometer cut points for predicting activity
  intensity in youth,''
\newblock {\em Medicine and Science in Sports and Exercise}, vol. 43, no. 7,
  pp. 1360--1368, 2011.

\bibitem{bouten1994assessment}
Carlijn Bouten, Klaas Westerterp, Maarten Verduin, and Jan Janssen,
\newblock ``Assessment of energy expenditure for physical activity using a
  triaxial accelerometer,''
\newblock {\em Medicine and Science in Sports and Exercise}, vol. 26, no. 12,
  pp. 1516--1523, 1994.

\bibitem{mathie2004classification}
MJ~Mathie, Branko~G Celler, Nigel~H Lovell, and ACF Coster,
\newblock ``Classification of basic daily movements using a triaxial
  accelerometer,''
\newblock {\em Medical and Biological Engineering and Computing}, vol. 42, no.
  5, pp. 679--687, 2004.

\bibitem{randell2000context}
Cliff Randell and Henk Muller,
\newblock ``Context awareness by analysing accelerometer data,''
\newblock in {\em Digest of Papers. Fourth International Symposium on Wearable
  Computers}. IEEE, 2000, pp. 175--176.

\bibitem{ravi2005activity}
Nishkam Ravi, Nikhil Dandekar, Preetham Mysore, and Michael~L Littman,
\newblock ``Activity recognition from accelerometer data,''
\newblock in {\em AAAI}, 2005, vol.~5, pp. 1541--1546.

\bibitem{gafurov2006biometric}
Davrondzhon Gafurov, Kirsi Helkala, and Torkjel S{\o}ndrol,
\newblock ``Biometric gait authentication using accelerometer sensor.,''
\newblock {\em JCP}, vol. 1, no. 7, pp. 51--59, 2006.

\bibitem{bourke2007evaluation}
AK~Bourke, JV~O’brien, and GM~Lyons,
\newblock ``Evaluation of a threshold-based tri-axial accelerometer fall
  detection algorithm,''
\newblock {\em Gait \& Posture}, vol. 26, no. 2, pp. 194--199, 2007.

\bibitem{bao2004activity}
Ling Bao and Stephen~S Intille,
\newblock ``Activity recognition from user-annotated acceleration data,''
\newblock in {\em International Conference on Pervasive Computing}. Springer,
  2004, pp. 1--17.

\bibitem{brezmes2009activity}
Tomas Brezmes, Juan-Luis Gorricho, and Josep Cotrina,
\newblock ``Activity recognition from accelerometer data on a mobile phone,''
\newblock in {\em International Work-Conference on Artificial Neural Networks}.
  Springer, 2009, pp. 796--799.

\bibitem{casale2011human}
Pierluigi Casale, Oriol Pujol, and Petia Radeva,
\newblock ``Human activity recognition from accelerometer data using a wearable
  device,''
\newblock in {\em Iberian Conference on Pattern Recognition and Image
  Analysis}. Springer, 2011, pp. 289--296.

\bibitem{kwapisz2011activity}
Jennifer~R Kwapisz, Gary~M Weiss, and Samuel~A Moore,
\newblock ``Activity recognition using cell phone accelerometers,''
\newblock {\em ACM SigKDD Explorations Newsletter}, vol. 12, no. 2, pp. 74--82,
  2011.

\bibitem{bayat2014study}
Akram Bayat, Marc Pomplun, and Duc~A Tran,
\newblock ``A study on human activity recognition using accelerometer data from
  smartphones,''
\newblock {\em Procedia Computer Science}, vol. 34, pp. 450--457, 2014.

\bibitem{siirtola2012recognizing}
Pekka Siirtola and Juha R{\"o}ning,
\newblock ``Recognizing human activities user-independently on smartphones
  based on accelerometer data,''
\newblock {\em International Journal of Interactive Multimedia and Artificial
  Intelligence}, vol. 1, no. 5, pp. 38--45, 2012.

\bibitem{hemminki2013accelerometer}
Samuli Hemminki, Petteri Nurmi, and Sasu Tarkoma,
\newblock ``Accelerometer-based transportation mode detection on smartphones,''
\newblock in {\em Proceedings of the 11th ACM Conference on Embedded Networked
  Sensor Systems}, 2013, pp. 1--14.

\bibitem{zhang2015recognizing}
Licheng Zhang, Xihong Wu, and Dingsheng Luo,
\newblock ``Recognizing human activities from raw accelerometer data using deep
  neural networks,''
\newblock in {\em 2015 IEEE 14th International Conference on Machine Learning
  and Applications (ICMLA)}. IEEE, 2015, pp. 865--870.

\bibitem{lee2017human}
Song-Mi Lee, Sang~Min Yoon, and Heeryon Cho,
\newblock ``Human activity recognition from accelerometer data using
  convolutional neural network,''
\newblock in {\em 2017 IEEE International Conference on Big Data and Smart
  Computing (BigComp)}. IEEE, 2017, pp. 131--134.

\bibitem{ignatov2018real}
Andrey Ignatov,
\newblock ``Real-time human activity recognition from accelerometer data using
  convolutional neural networks,''
\newblock {\em Applied Soft Computing}, vol. 62, pp. 915--922, 2018.

\bibitem{hassan2018robust}
Mohammed~Mehedi Hassan, Md~Zia Uddin, Amr Mohamed, and Ahmad Almogren,
\newblock ``A robust human activity recognition system using smartphone sensors
  and deep learning,''
\newblock {\em Future Generation Computer Systems}, vol. 81, pp. 307--313,
  2018.

\bibitem{wittenburg2006elan}
Peter Wittenburg, Hennie Brugman, Albert Russel, Alex Klassmann, and Han
  Sloetjes,
\newblock ``{ELAN}: a professional framework for multimodality research,''
\newblock in {\em 5th International Conference on Language Resources and
  Evaluation (LREC 2006)}, 2006.

\bibitem{pahar2021automatic}
Madhurananda Pahar, Igor Miranda, Andreas Diacon, and Thomas Niesler,
\newblock ``Automatic non-invasive cough detection based on accelerometer and
  audio signals,''
\newblock {\em arXiv preprint arXiv:2109.00103}, 2021.

\bibitem{takahashi2016acoustic}
Gen Takahashi, Takeshi Yamada, Shoji Makino, and Nobutaka Ono,
\newblock ``Acoustic scene classification using deep neural network and
  frame-concatenated acoustic feature,''
\newblock {\em Detection and Classification of Acoustic Scenes and Events},
  2016.

\bibitem{van2007experimental}
Jason Van~Hulse, Taghi~M Khoshgoftaar, and Amri Napolitano,
\newblock ``Experimental perspectives on learning from imbalanced data,''
\newblock in {\em Proceedings of the 24th International Conference on Machine
  Learning}, 2007, pp. 935--942.

\bibitem{krawczyk2016learning}
Bartosz Krawczyk,
\newblock ``Learning from imbalanced data: open challenges and future
  directions,''
\newblock {\em Progress in Artificial Intelligence}, vol. 5, no. 4, pp.
  221--232, 2016.

\bibitem{chawla2002smote}
Nitesh~V Chawla, Kevin~W Bowyer, Lawrence~O Hall, and W~Philip Kegelmeyer,
\newblock ``{SMOTE}: synthetic minority over-sampling technique,''
\newblock {\em Journal of {A}rtificial {I}ntelligence {R}esearch}, vol. 16, pp.
  321--357, 2002.

\bibitem{lemaitre2017imbalanced}
Guillaume Lema{\^\i}tre, Fernando Nogueira, and Christos~K Aridas,
\newblock ``Imbalanced-learn: A {P}ython toolbox to tackle the curse of
  imbalanced datasets in machine learning,''
\newblock {\em The Journal of Machine Learning Research}, vol. 18, no. 1, pp.
  559--563, 2017.

\bibitem{windmon2018tussiswatch}
Anthony Windmon, Mona Minakshi, Pratool Bharti, Sriram Chellappan, Marcia
  Johansson, Bradlee~A Jenkins, and Ponrathi~R Athilingam,
\newblock ``Tussiswatch: A smart-phone system to identify cough episodes as
  early symptoms of chronic obstructive pulmonary disease and congestive heart
  failure,''
\newblock {\em IEEE Journal of Biomedical and Health Informatics}, vol. 23, no.
  4, pp. 1566--1573, 2018.

\bibitem{BlagusSMOTE}
R.~Blagus and L~Lusa,
\newblock ``{SMOTE} for high-dimensional class-imbalanced data,''
\newblock {\em BMC Bioinformatics}, vol. 14, pp. 106, 2013.

\bibitem{han2005borderline}
Hui Han, Wen-Yuan Wang, and Bing-Huan Mao,
\newblock ``Borderline-{SMOTE}: a new over-sampling method in imbalanced data
  sets learning,''
\newblock in {\em International {C}onference on {I}ntelligent {C}omputing}.
  Springer, 2005, pp. 878--887.

\bibitem{nguyen2011borderline}
Hien~M Nguyen, Eric~W Cooper, and Katsuari Kamei,
\newblock ``Borderline over-sampling for imbalanced data classification,''
\newblock {\em International Journal of Knowledge Engineering and Soft Data
  Paradigms}, vol. 3, no. 1, pp. 4--21, 2011.

\bibitem{he2008adasyn}
Haibo He, Yang Bai, Edwardo~A Garcia, and Shutao Li,
\newblock ``Adasyn: Adaptive synthetic sampling approach for imbalanced
  learning,''
\newblock in {\em 2008 IEEE {I}nternational {J}oint {C}onference on {N}eural
  {N}etworks (IEEE {W}orld {C}ongress on {C}omputational {I}ntelligence)}.
  IEEE, 2008, pp. 1322--1328.

\bibitem{pan2012investigation}
Jia Pan, Cong Liu, Zhiguo Wang, Yu~Hu, and Hui Jiang,
\newblock ``Investigation of deep neural networks ({DNN}) for large vocabulary
  continuous speech recognition: Why {DNN} surpasses {GMM}s in acoustic
  modeling,''
\newblock in {\em 2012 8th International Symposium on Chinese Spoken Language
  Processing}. IEEE, 2012, pp. 301--305.

\bibitem{christodoulou2019systematic}
Evangelia Christodoulou, Jie Ma, Gary~S Collins, Ewout~W Steyerberg, Jan~Y
  Verbakel, and Ben Van~Calster,
\newblock ``A systematic review shows no performance benefit of machine
  learning over logistic regression for clinical prediction models,''
\newblock {\em Journal of Clinical Epidemiology}, vol. 110, pp. 12--22, 2019.

\bibitem{botha2018detection}
GHR Botha, G~Theron, RM~Warren, M~Klopper, K~Dheda, PD~Van~Helden, and
  TR~Niesler,
\newblock ``Detection of tuberculosis by automatic cough sound analysis,''
\newblock {\em Physiological Measurement}, vol. 39, no. 4, pp. 045005, 2018.

\bibitem{le1992ridge}
Saskia Le~Cessie and Johannes~C Van~Houwelingen,
\newblock ``Ridge estimators in logistic regression,''
\newblock {\em Journal of the Royal Statistical Society: Series C (Applied
  Statistics)}, vol. 41, no. 1, pp. 191--201, 1992.

\bibitem{tsuruoka2009stochastic}
Yoshimasa Tsuruoka, Jun’ichi Tsujii, and Sophia Ananiadou,
\newblock ``Stochastic gradient descent training for l1-regularized log-linear
  models with cumulative penalty,''
\newblock in {\em Proceedings of the Joint Conference of the 47th Annual
  Meeting of the ACL and the 4th International Joint Conference on Natural
  Language Processing of the AFNLP}, 2009, pp. 477--485.

\bibitem{yamashita2003interior}
Hiroshi Yamashita and Hiroshi Yabe,
\newblock ``An interior point method with a primal-dual quadratic barrier
  penalty function for nonlinear optimization,''
\newblock {\em SIAM Journal on Optimization}, vol. 14, no. 2, pp. 479--499,
  2003.

\bibitem{taud2018multilayer}
H~Taud and JF~Mas,
\newblock ``Multilayer perceptron ({MLP}),''
\newblock in {\em Geomatic Approaches for Modeling Land Change Scenarios}, pp.
  451--455. Springer, 2018.

\bibitem{sarangi2016design}
Lokanath Sarangi, Mihir~Narayan Mohanty, and Srikanta Pattanayak,
\newblock ``Design of {MLP} based model for analysis of patient suffering from
  influenza,''
\newblock {\em Procedia Computer Science}, vol. 92, pp. 396--403, 2016.

\bibitem{tracey2011cough}
Brian~H Tracey, Germ{\'a}n Comina, Sandra Larson, Marjory Bravard, Jos{\'e}~W
  L{\'o}pez, and Robert~H Gilman,
\newblock ``Cough detection algorithm for monitoring patient recovery from
  pulmonary tuberculosis,''
\newblock in {\em 2011 Annual International Conference of the IEEE Engineering
  in Medicine and Biology Society}. IEEE, 2011, pp. 6017--6020.

\bibitem{liu2014cough}
Jia-Ming Liu, Mingyu You, Zheng Wang, Guo-Zheng Li, Xianghuai Xu, and Zhongmin
  Qiu,
\newblock ``Cough detection using deep neural networks,''
\newblock in {\em 2014 IEEE International Conference on Bioinformatics and
  Biomedicine (BIBM)}. IEEE, 2014, pp. 560--563.

\bibitem{amoh2015deepcough}
Justice Amoh and Kofi Odame,
\newblock ``Deepcough: A deep convolutional neural network in a wearable cough
  detection system,''
\newblock in {\em 2015 IEEE Biomedical Circuits and Systems Conference
  (BioCAS)}. IEEE, 2015, pp. 1--4.

\bibitem{krizhevsky2017imagenet}
Alex Krizhevsky, Ilya Sutskever, and Geoffrey~E Hinton,
\newblock ``Imagenet classification with deep convolutional neural networks,''
\newblock {\em Communications of the ACM}, vol. 60, no. 6, pp. 84--90, 2017.

\bibitem{pahar2020covid}
Madhurananda Pahar, Marisa Klopper, Robin Warren, and Thomas Niesler,
\newblock ``{COVID}-19 cough classification using machine learning and global
  smartphone recordings,''
\newblock {\em Computers in Biology and Medicine}, vol. 135, pp. 104572, 2021.

\bibitem{pahar_coding_2020}
Madhurananda Pahar and Leslie~S Smith,
\newblock ``Coding and {D}ecoding {S}peech using a {B}iologically {I}nspired
  {C}oding {S}ystem,''
\newblock in {\em 2020 IEEE Symposium Series on Computational Intelligence
  (SSCI)}. IEEE, 2020, pp. 3025--3032.

\bibitem{pahar2021tb}
Madhurananda Pahar, Marisa Klopper, Byron Reeve, Rob Warren, Grant Theron, and
  Thomas Niesler,
\newblock ``Automatic cough classification for tuberculosis screening in a
  real-world environment,''
\newblock {\em Physiological Measurement}, vol. 42, no. 10, pp. 105014, oct
  2021.

\bibitem{lawrence1997face}
Steve Lawrence, C~Lee Giles, Ah~Chung Tsoi, and Andrew~D Back,
\newblock ``Face recognition: {A} convolutional neural-network approach,''
\newblock {\em IEEE Transactions on Neural Networks}, vol. 8, no. 1, pp.
  98--113, 1997.

\bibitem{albawi2017understanding}
Saad Albawi, Tareq~Abed Mohammed, and Saad Al-Zawi,
\newblock ``Understanding of a convolutional neural network,''
\newblock in {\em 2017 International Conference on Engineering and Technology
  (ICET)}. IEEE, 2017, pp. 1--6.

\bibitem{qi2017comparison}
Xingqun Qi, Tianhui Wang, and Jiaming Liu,
\newblock ``Comparison of support vector machine and softmax classifiers in
  computer vision,''
\newblock in {\em 2017 Second International Conference on Mechanical, Control
  and Computer Engineering (ICMCCE)}. IEEE, 2017, pp. 151--155.

\bibitem{hochreiter1997long}
Sepp Hochreiter and J{\"u}rgen Schmidhuber,
\newblock ``Long short-term memory,''
\newblock {\em Neural Computation}, vol. 9, no. 8, pp. 1735--1780, 1997.

\bibitem{miranda2019comparative}
Igor~DS Miranda, Andreas~H Diacon, and Thomas~R Niesler,
\newblock ``A comparative study of features for acoustic cough detection using
  deep architectures,''
\newblock in {\em 2019 41st Annual International Conference of the IEEE
  Engineering in Medicine and Biology Society (EMBC)}. IEEE, 2019, pp.
  2601--2605.

\bibitem{marchi2015non}
Erik Marchi, Fabio Vesperini, Felix Weninger, Florian Eyben, Stefano Squartini,
  and Bj{\"o}rn Schuller,
\newblock ``Non-linear prediction with {LSTM} recurrent neural networks for
  acoustic novelty detection,''
\newblock in {\em 2015 International Joint Conference on Neural Networks
  (IJCNN)}. IEEE, 2015, pp. 1--7.

\bibitem{amoh2016deep}
Justice Amoh and Kofi Odame,
\newblock ``Deep neural networks for identifying cough sounds,''
\newblock {\em IEEE Transactions on Biomedical Circuits and Systems}, vol. 10,
  no. 5, pp. 1003--1011, 2016.

\bibitem{sherstinsky2020fundamentals}
Alex Sherstinsky,
\newblock ``Fundamentals of recurrent neural network ({RNN}) and long
  short-term memory ({LSTM}) network,''
\newblock {\em Physica D: Nonlinear Phenomena}, vol. 404, pp. 132306, 2020.

\bibitem{Sammut2010}
``Leave-one-out cross-validation,''
\newblock in {\em Encyclopedia of Machine Learning}, Claude Sammut and
  Geoffrey~I. Webb, Eds., Boston, MA, 2010, pp. 600--601, Springer US.

\bibitem{wong2015performance}
Tzu-Tsung Wong,
\newblock ``Performance evaluation of classification algorithms by k-fold and
  leave-one-out cross validation,''
\newblock {\em Pattern Recognition}, vol. 48, no. 9, pp. 2839--2846, 2015.

\bibitem{diacon2012time}
AH~Diacon, JS~Maritz, A~Venter, PD~Van~Helden, R~Dawson, and PR~Donald,
\newblock ``Time to liquid culture positivity can substitute for colony
  counting on agar plates in early bactericidal activity studies of
  antituberculosis agents,''
\newblock {\em Clinical Microbiology and Infection}, vol. 18, no. 7, pp.
  711--717, 2012.

\bibitem{he2016deep}
Kaiming He, Xiangyu Zhang, Shaoqing Ren, and Jian Sun,
\newblock ``Deep residual learning for image recognition,''
\newblock in {\em Proceedings of the IEEE {C}onference on {C}omputer {V}ision
  and {P}attern {R}ecognition}, 2016, pp. 770--778.

\end{thebibliography}

\end{document}